\documentclass{article}

\usepackage{microtype}
\usepackage{graphicx}
\usepackage{subcaption}
\usepackage{booktabs}

\usepackage[preprint]{neurips_2026}

\usepackage[utf8]{inputenc} % allow utf-8 input
\usepackage[T1]{fontenc}    % use 8-bit T1 fonts
\usepackage[colorlinks=true,
            linkcolor=red,
            citecolor=blue,
            filecolor=blue,
            urlcolor=blue]{hyperref}
\usepackage{url}            % simple URL typesetting
\usepackage{booktabs}       % professional-quality tables
\usepackage{amsfonts}       % blackboard math symbols
\usepackage{nicefrac}       % compact symbols for 1/2, etc.
\usepackage{microtype}      % microtypography
\usepackage{xcolor}         % colors
\usepackage{amsmath}
\usepackage{amssymb}

\usepackage{float}
\usepackage{xcolor}
\usepackage{verbatim}
\usepackage{wrapfig}
\usepackage{siunitx}
\newtheorem{lemma}{Lemma}

\title{SimReg: Achieving Higher Performance in the Pretraining via Embedding Similarity Regularization}

\author{
\textbf{Yan Sun}$^{1}$,
\textbf{Guoxia Wang}$^{1}$,
\textbf{Jinle Zeng}$^{1}$,
\textbf{JiaBin Yang}$^{1}$,
\textbf{Shuai Li}$^{1}$ \\
\textbf{Li Shen}$^{3}$,
\textbf{Dacheng Tao}$^{4}$,
\textbf{DianHai Yu}$^{1}$,
\textbf{Haifeng Wang}$^{1}$ \\
$^{1}$Baidu Inc. \ \
$^{2}$Sun Yat-sen University \ \
$^{3}$Nanyang Technological University \\
\texttt{\{sunyan25,wangguoxia,zengjinle,yangjiabin01,yudianhai,wanghaifeng\}@baidu.com} \\
\texttt{lishuai\_math@163.com, mathshenli@gmail.com, dacheng.tao@ntu.edu.sg}
}

\begin{document}
\maketitle

\begin{abstract}
  Pretraining large language models~(LLMs) with next-token prediction has led to remarkable advances, yet the context-dependent nature of token embeddings in such models results in high intra-class variance and inter-class similarity, thus hindering the efficiency of representation learning. While similarity-based regularization has demonstrated benefit in supervised fine-tuning and classification tasks, its application and efficacy in large-scale LLM pretraining remains underexplored. In this work, we propose the \textsc{SimReg}, an embedding similarity regularization loss that explicitly encourages token representations with the same ground-truth label within each sequence to be more similar, while enforcing separation from different-label tokens via a contrastive loss. Our analysis reveals that this mechanism introduces gains by enlarging multi-classification margins, thereby enabling more efficient classification. Extensive experiments across dense and Mixture-of-Experts~(MoE) architectures demonstrate that \textsc{SimReg} consistently accelerates training convergence by over $30\%$ and improves average zero-shot downstream performance by over $1\%$ across standard benchmarks. Further ablation studies and analyses offer practical insights into hyperparameter tuning and loss effectiveness.
\end{abstract}
\section{Introduction}
LLMs have emerged as a cornerstone of modern artificial intelligence and have demonstrated remarkable capabilities across a wide range of domains such as natural language understanding~\citep{radford2019language}, reasoning~\citep{wei2022chain}, and multimodal interaction~\citep{lin2025survey}. While LLMs are advancing along diverse directions, they all fundamentally share a consistent underlying principle, i.e., next-token prediction. The essential mechanism of LLMs is to predict the categorical distribution of the next token from the embeddings of the prior context, which can also be viewed as a classification problem defined over the combined representations of the preceding context. By leveraging enormous model parameters and vast training data, it exhibits exceptional generalization capability, introduces novel solutions in diverse research domains, and further drives the adoption of a wide range of applications~\citep{topsakal2023creating} with growing challenges in efficiency~\citep{shen2024efficient}. Both data-specific~\citep{fan2025joint,deng2026less} and weight-specific~\citep{li2024fast,sun2025maskpro,lin2025awq} approaches have attracted considerable research interest.

Unlike conventional classification, language model prediction does not rely on a stable object strictly tied to its label. In image classification, for instance, a cat image is consistently associated with its label, leading to highly consistent embeddings within the same class. In contrast, on language tasks, the representation used to predict a token is composed of diverse contextual features, many unrelated to the label itself. As a result, embeddings predicting the same token can vary significantly. For example, the representations for “walls” in “The cat jumps over walls” and “A child paints near walls” originate from entirely different contexts, making the classification process more challenging.

Recent advances in consistency learning for finetuning language models shed light on potential solutions to this challenge~\citep{Huang2021TokenLevelSC,gunel2021supervised,yin2023consistency}. However, this line of research has not yet been extended to pretraining and has not been widely adopted in the large-scale pretraining practices. Post-training is typically performed with a small learning rate and limited datasets, which makes it difficult to significantly modify the geometric structure of the learned parameters. These insights motivate us to further extend this approach to large-scale pretraining.

\begin{figure*}[t]
\centering
% \vskip -0.1in
\includegraphics[width=0.9\textwidth]{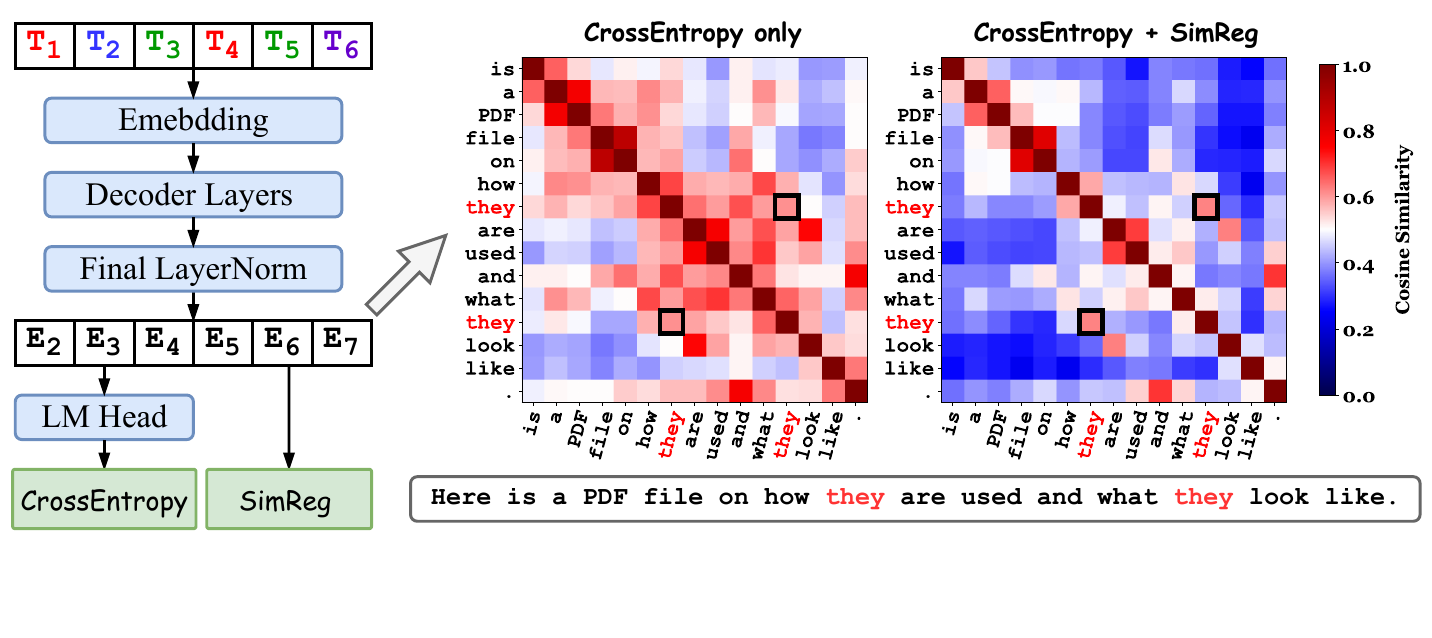}
\vskip -0.3in
\caption{(left)~Workflow of the \textsc{SimReg} loss. (Right)~We compare the cosine similarity of token embeddings in a sample on the LLaMA-7B model trained via ``CrossEntropy only" and ``CrossEntropy+\textsc{SimReg}". Using ``CrossEntropy only" fails to enforce sufficient separability among token features, whose cosine values of all token pairs exceed $0.5$. With the introduction of \textsc{SimReg}, feature separability is generally enhanced~(averaged cosine value is reduced by at least $0.1$), thereby providing stronger support for classification training. More results are stated in Appendix~\ref{ap:visualization}.}
\label{fg:pipeline}
\end{figure*}

In this work, we show that large-scale pretraining with cross-entropy alone fails to impose strong consistency on token embeddings. To address this, we then add a consistency regularization term, \textsc{SimReg}, to strengthen the representational capacity of large models during pretraining. For each token in a sequence, all tokens are partitioned into positive and negative groups. The objective penalizes the similarity across groups, which pulls embeddings toward same-class samples and pushes them away from different-class samples. To ensure valid contrastive pairs for every token, \textsc{SimReg} introduces self-sample similarity in each positive group and further computes the loss with group-level rather than sample-level averaging, which balances the contributions of different tokens, which allows it to preserve a high level of stability over the long pretraining runs. We also provide a thorough theoretical understanding to explain how it contributes to improving cross-entropy loss. Extensive evaluations are conducted on both dense and MoE models, including LLaMA-350M, 1.3B, 3B, 7B~\citep{touvron2023llama}, and Mixtral-8$\times$1B~\citep{jiang2024mixtral}. The \textsc{SimReg} loss can consistently accelerates convergence by over 30\% in pretraining. When training with over 52B tokens, it can yield an improvement of more than 1\% in average performance across downstream general tasks. We investigate the hyperparameter sensitivity of \textsc{SimReg} and find that it maintains a wide range of applicability. We summarize our main contributions as follows:
\begin{itemize}
    \item We explore the advantages of employing consistency regularization in large-scale pretraining tasks and propose a series of improvements to address the training instabilities of existing methods, thereby enabling stable performance gains throughout long-term pretraining.
    \item We provide a detailed theoretical analysis of the benefits of the \textsc{SimReg} loss for the cross-entropy loss, and how it improves the multi-classification margins.
    \item We conduct extensive experiments to validate its substantial improvements for pretraining tasks, achieving an average training acceleration of over 30\% and yielding over 1\% gains on downstream tasks, and state detailed empirical insights for the community.
\end{itemize}

\section{Related Work}

\textbf{Contrastive learning}. The systematic exploration of feature similarity constraints in machine learning can be traced back to their early development in computer vision~(CV) tasks and contrastive learning~\citep{oord2018representation,khosla2020supervised}. They enhance the training of baseline classification models by constructing virtual data pairs and incorporating additional supervised loss signals, which helped the models extract more discriminative features. It is typically employed to counteract noise perturbations at the input level, thereby improving generalization ability~\citep{geng2021context,shi2022simple,huang2022contrastive,zhou2024contrastive,wang2024generated}. Generally, a data pair is constructed from a raw sample and its perturbed counterpart, and the model is trained to minimize their representation similarity. Subsequently, supervised contrastive learning has been extended to incorporate class information. By leveraging available labels to construct class-consistent data pairs, the model is trained not only to pull together samples from the same class but also to push apart samples from different classes~\citep{wang2021understanding,wen2021toward,ye2022unsupervised,denize2023similarity}. Recent studies have revealed that contrastive learning can also achieve more efficient feature extraction across tasks and data originating from different domains~\citep{verma2021towards,wang2022cross,xie2022contrastive,azuma2023adversarial}. In multimodal large model training, this learning paradigm is often employed to align the mapping of knowledge across domains and to capture the representation capacity of the same knowledge under different modalities~\citep{yuan2021multimodal,mai2022hybrid,liu2024contrastive,sun2024nodule}. In summary, contrastive learning offers an efficient and general paradigm for representation learning to the machine learning community.

\textbf{Embedding Consistency in LLMs.} The study of feature similarity has also been considered as compositional generalization~\citep{lake2019compositional,wiedemer2023compositional} and embedding consistency regularization~\citep{yin2023consistency}. \citet{gao2021simcse} learn the sentence embeddings and achieve higher generalization efficiency. Then it is widely expanded to the token-level~\citep{gao2023empirical,wang2023going}, word-level~\citep{kenter2015short,antoniak2018evaluating}, context-level~\citep{laskar2020contextualized}. Most of these tasks have primarily focused on small-scale or fine-tuning settings. As the cornerstone of modern language models, the next-token prediction paradigm has been widely applied across various downstream tasks~\citep{li2024mechanics,chen2024next}. Recent research has further investigated the similarity and dispersion of token embeddings, which highlights the separability of embeddings to be a key direction~\citep{de2023class,tao2024llms,hu2024enhancing}.
\section{Problem Setup and Methodology}

In this section, we introduce how \textsc{SimReg} can be incorporated into the pretraining of LLMs and explain why it helps improve performance. Before proceeding, we formalize the overall pretraining setup of LLMs and introduce the notations used throughout the subsequent analysis.

\textbf{General Pretraining.} Before introducing the training framework, we first define the notation in this work. We consider the progress of LLM pretraining as learning the optimal weight $\mathbf{w}$ by minimizing the cross-entropy loss $\ell$ under a general distribution $\mathcal{D}$. We decompose the model into two cascaded functions $f_P \circ f_E$, where $f_P$ (the logits generation module) is parameterized by $\mathbf{w}_P$ and $f_E$ (the embedding generation module) is parameterized by $\mathbf{w}_E$, with the overall parameters denoted as $\mathbf{w}=\left[\mathbf{w}_P, \mathbf{w}_E\right]$. Based on this decomposition, the general pretraining objective of language models can then be formally formulated as:
\begin{equation}
\label{eq:ce}
    \min_\mathbf{w} \mathbb{E}_{\left(\mathbf{x}_i,y_i\right)\sim\mathcal{D}} \left[\ell\left(f_P \circ f_E\left(\mathbf{x}_i\right), y_i\right)\right],
\end{equation}
where $\left(\mathbf{x}_i,y_i\right)$ is the $\left(\text{data}, \text{label}\right)$ pair sampled from the distribution $\mathcal{D}$. Here, the choice of $f_E$ and $f_P$ is entirely flexible, meaning that the \textsc{SimReg} loss can in principle be applied to any valid token embedding across the network. We further explore the optimal placement of this component in subsequent experiments of Sec.~\ref{exp:position}.

Cross-entropy loss serves as the fundamental training objective in language modeling. It measures the discrepancy between the predicted token distribution and the ground-truth one-hot distribution, thereby guiding the model to maximize the likelihood of the correct next token. The models typically employ large-scale feature extractors to obtain separable representations. By denoting the token embedding as $\mathbf{e}_i=f_E(\mathbf{x}_i)$ and corresponding logits as $\mathbf{z}_i=f_P(\mathbf{e}_i)$, the population risk of sample-wise cross-entropy loss is:
\begin{equation}
\label{eq:cr_loss}
    L^{\text{ce}} = \frac{1}{n}\sum_i \left(- \mathbf{z}_{i,y_i} + \log\left(\sum_j \exp\left({\mathbf{z}_{i,j}}\right)\right)\right).
\end{equation}

\begin{figure*}[t]
\centering
\subfloat[Imbalanced Token Frequency in Pretraining.]{
  \includegraphics[width=0.48\textwidth]{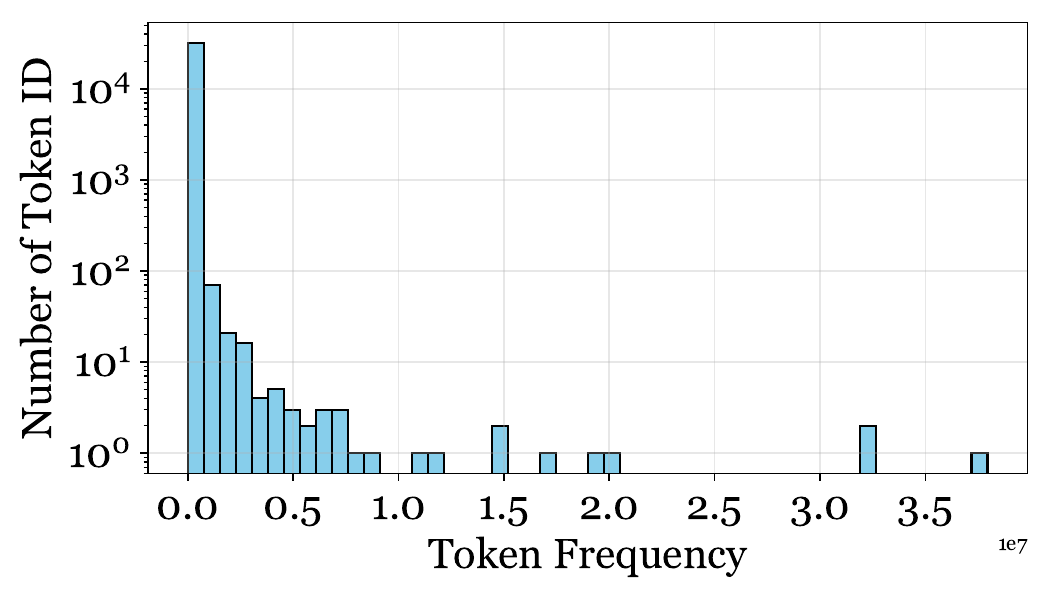}
}
\hfill
\subfloat[Limited Contrastive Similarity in Pretraining.]{
  \includegraphics[width=0.48\textwidth]{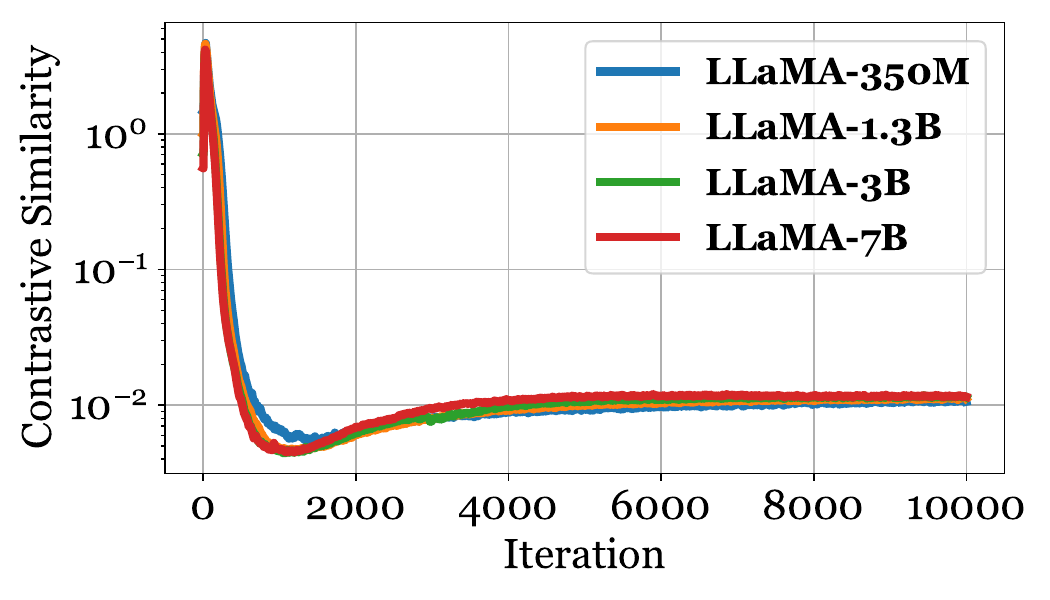}
}
% \vskip -0.05in
\caption{
(a) We analyze the token ID distribution over 1B training samples from the C4 dataset and find that only about 2\% of tokens occur with extremely high frequency, resulting in a pronounced long-tail effect in the classification data.
(b) We observe that the contrastive similarity loss of embeddings does not continue to decrease after reaching a basic threshold and then the feature similarity is no longer further optimized. \textbf{Simply increasing the model size does not improve this performance.}
}
\label{fg:observation:loss}
% \vskip -0.1in
\end{figure*}

Generally, larger separability can enhance the distinction between different samples, leading to more robust and discriminative representations. Although Eq.~(\ref{eq:cr_loss}) averages over samples, the unique characteristics of language tasks introduce a challenge: \textit{the distribution of words (tokens) is highly imbalanced}, which causes frequent tokens to dominate the loss while rare but informative ones contribute disproportionately little, yielding a heavy long-tail dataset. When training classification tasks on such dataset, the inter-class margin is greatly influenced by the number of samples per class. As shown in Figure~\ref{fg:observation:loss}~(a), we empirical investigate the token distribution of C4 dataset and the behavior of contrastive similarity. \textbf{A primary challenge we investigate in the LLM pretraining is:}
\begin{center}
\textit{cross-entropy \textbf{stops} driving stronger representation learning after a basic separability level of tokens.}
\end{center}
Figure~\ref{fg:observation:loss}.(b) indicates that during the early stage of cross-entropy training, the model rapidly constrains the contrastive similarity of embeddings. However, once the contrastive diversity becomes sufficient to sustain classification training, the model no longer enforces heterogeneity among token embeddings. Subsequently, even though the cross-entropy loss continues to decrease, the contrastive similarity exhibits little further change. Another interesting phenomenon we observe is that, even as the model depth increases and the embedding dimension grows, the supervision of token embedding contrastive similarity under cross-entropy remains nearly at the same level. This limits the potential for further improvement in classification tasks, while also motivating us to impose the contrastive similarity.

\textbf{Embedding Similarity Regularization.} 
%Previous work has already explored this direction in small-scale training and fine-tuning tasks~\citep{gao2021simcse,li2024mechanics}. 
Here, we introduce the generalized form of our similarity regularization. For each token $\mathbf{x}$, its embedding can be denoted by $\mathbf{e}=f_E(\mathbf{x})$. For each data sample $\left(\mathbf{x}_i,y_i\right)$, we can define a positive embedding set $\mathcal{P}_{i}=\left\{k:y_k=y_i\right\}$ and a negative embedding set $\mathcal{N}_i=\left\{k:y_k\neq y_i\right\}$. The consistency loss aims to minimize the distance between embeddings of positive pairs, while simultaneously maximizing the separation between negative pairs:
\begin{equation}
\label{eq:general simreg}
    L_i^{\text{sr}} \triangleq \log\sum_{j\in\mathcal{N}_i}\phi_{i,j}-\log\sum_{j\in\mathcal{P}_i}\phi_{i,j},
\end{equation}
where $L_i^{\text{sr}}$ is the similarity loss of $i$-th token. $\phi$ denotes a similarity function between two embeddings. We explore two primary forms: the exponential of the inner-product $\left\langle \mathbf{e}_i, \mathbf{e}_j\right\rangle$ and that of the cosine similarity $\frac{\left\langle \mathbf{e}_i, \mathbf{e}_j\right\rangle}{\Vert \mathbf{e}_i\Vert\cdot\Vert\mathbf{e}_j \Vert}$. Both similarity measures provide effective supervision for feature similarity, yet their applicable scenarios differ. It often yields stronger statistical constraints, thereby enforcing supervision on both geometric structure and feature norms. However, this advantage may also introduce ambiguity: for instance, when a embedding has an abnormally large norm, the inner-product value becomes dominated by the magnitude, rendering the loss function almost insensitive to angular differences. In such cases, the optimization may overly rely on vector norms while neglecting the discriminative power of directional alignment. Therefore, for numerical stability, we adopt cosine similarity as the similarity measure in Eq.~(\ref{eq:general simreg}) and introduce a constant temperature coefficient $\tau$ to adjust the sharpness of the distribution. Since words in natural language are inherently distributed in an imbalanced manner, a sequence may contain only a single occurrence of a particular token type, we add the self-similarity to $\mathcal{P}_i$ to enforce that there exists at least a positive data pair. Moreover, to ensure that the regularization loss is non-negative, we introduce the {\ttfamily softplus} function to further scale it. Therefore, the final form of the loss is computed as $L_i=L_i^\text{ce}+\lambda\cdot\text{softplus}\left(L_i^\text{sr}\right)$. The entire optimization process involves two hyperparameters $\tau$ and $\lambda$. We discuss them in Sec.~\ref{exp:hyperparameter sensitivity}.

\textbf{Chunk-wise \textsc{SimReg} for Sequence Parallelism.} The computation of \textsc{SimReg} is centered on the embedding of each token within a sequence sample, whose complexity is $\mathcal{O}(n^2)$. Its computation can naturally support parallelization strategies like data parallelism~(DP), tensor parallelism~(TP), and pipeline parallelism~(PP). However, during long-text training, sequence parallelism~(SP) splits the data of each sequence across different nodes for training, which introduces additional redundant communications. To alleviate this issue, we divide \textsc{SimReg} into $b$ chunks, where every $\frac{n}{b}$ tokens form a chunk to compute the \textsc{SimReg} loss internally. The losses across different nodes are then weighted according to the ratio of positive and negative samples, while the overall computational complexity is reduced to $\mathcal{O}(n^2/b)$. Moreover, we further point out that there exists a fundamental trade-off between the strength of supervision and the expressive capacity of feature representations with respect to the choice of chunk size. When a chunk contains a larger number of tokens, the estimation of \textsc{SimReg} becomes more accurate. However, its constraining power on each individual token is weakened, as the loss must balance relationships among a larger set of tokens. Therefore, selecting an appropriate chunk size is of critical importance, which is empirically explored in Sec.~\ref{exp:emp_studies}.
\section{Theoretical Analysis}
In this section, we demonstrate how \textsc{SimReg} improves classification margins. All proofs are provided in Appendix~\ref{ap:proof}. We first introduce classification margins, i.e., $m_i=\mathbf{z}_{i,y_i}-\max_{j\neq y_i}\mathbf{z}_{i,j}$, which is the gap between the top two logits. Then the cross entropy loss can be upper bounded by:
\begin{equation}
\begin{split}
    \ell_i=\log\left(1+\sum_j\exp\left(\mathbf{z}_{i,j}-\mathbf{z}_{i,y_i}\right)\right) \leq C\exp\left(-m_i\right),
\end{split}
\end{equation}
where $C$ is the number of total classes. The above formula explicitly characterizes the relationship between the classification margin and the training loss, and enlarging the margin $m$ leads to a further reduction in the loss. Our supervision on the embedding variable $\mathbf{e}$ is propagated through a function $f_P\left(\cdot\right)$ to the logits $\mathbf{z}$ used for the classification with cross-entropy, i.e., $\mathbf{z}=f_P\left(\mathbf{e}\right)$. This mapping can take the form of a simple linear projection (e.g., the LM head) or several intermediate layers of the LLM. Without loss of generality, we assume it to be a general smooth and non-convex function with smoothness coefficient $L_P$. Thus, we consider the margin. By defining $\mathbb{I}$ as the standard basis vector where $\mathbb{I}_j$ means $1$ in the $j$-th coordinate and $0$ elsewhere, we measure the pair-wise gap in logits by $g_{y_k,j}(\mathbf{e}_k)=\left(\mathbb{I}_{y_k}-\mathbb{I}_j\right)^\top f_P(\mathbf{e}_k)$, which also holds smoothness and non-convexity. Furthermore, we can transfer the smoothness by: $\vert g_{y_i,j}(\mathbf{e}_p) - g_{y_i,j}(\mathbf{e}_q)\vert\leq\Vert \mathbb{I}_{y_i} - \mathbb{I}_j \Vert\Vert f_P(\mathbf{e}_p) - f_P(\mathbf{e}_q) \Vert\leq\sqrt{2}L_P\Vert\mathbf{e}_p-\mathbf{e}_q\Vert$. To investigate their relationships, we have the following lemma.
\begin{lemma}
    For each token $\mathbf{x}_i$ where its embedding is $\mathbf{e}_i=f_E\left(\mathbf{x}_i\right)$, we further define a weighted center of the embedding in the original space, where the positive and negative centers are $\overline{\mathbf{e}}_k^+=\frac{\sum_{i\in\mathcal{P}_k}\alpha_{k,i}\mathbf{e}_i}{\sum_{i\in\mathcal{P}_k}\alpha_{k,i}}$ and $\overline{\mathbf{e}}_k^-=\frac{\sum_{i\in\mathcal{N}_k}\alpha_{k,i}\mathbf{e}_i}{\sum_{i\in\mathcal{N}_k}\alpha_{k,i}}$ where $\alpha_{k,i}\propto\exp\left(\mathbf{e}_k^\top\mathbf{e}_i\right)$. Then we have the averaged group margins are $\overline{m}_k^+=\min_{c\neq y_k}g_{y_k,c}(\overline{\mathbf{e}}_k^+)$ and $\overline{m}_k^-=\min_{c\neq y_k}g_{y_k,c}(\overline{\mathbf{e}}_k^-)$. Therefore, the classification margin bound of each token $m_k$ is the Central–Eccentric lower bound within the group margin:
    \begin{equation}
        \overline{m}_k^+ - \sqrt{2}L_P \Vert\mathbf{e}_k-\overline{\mathbf{e}}_k^+\Vert\leq m_k\leq  \overline{m}_k^- 
        + \sqrt{2}L_P\Vert\mathbf{e}_k-\overline{\mathbf{e}}_k^-\Vert.
    \end{equation}
\end{lemma}
Intuitively, $\overline{m}$ can be regarded as an idealized margin, obtained by evaluating the logit of the correct class at the positive center and that of the strongest competing class at the negative center. 
%Compared with evaluating directly at the embedding $\mathbf{e}$, this formulation is more effective because the centers are smoothed from many neighboring samples with exponential cosine weighting and this makes the margin a clearer and more stable measure of intra-class alignment and inter-class separation.
A key point is that it separates the upper and lower bounds of the classification margin for each individual sample, showing that the lower bound is influenced by the distance to positive samples $\Vert\mathbf{e}_k-\overline{\mathbf{e}}_k^+\Vert$ and $\overline{m}_k^+$, while the upper bound is determined by the distance to negative samples $\Vert\mathbf{e}_k-\overline{\mathbf{e}}_k^-\Vert$ and $\overline{m}_k^-$. Thus we have:
\begin{itemize}
    \item The dynamics of the central distance of the positive set would decrease: $\frac{d}{dt}\Vert \mathbf{e}_k-\overline{\mathbf{e}}_k^+\Vert^2\leq 0$;
    \item The classification margin at the positive center would increase: there exists a positive constant $\delta$ that
    \begin{equation}
        g_{y_k,j}(\mathbf{e}_k^+ + \epsilon_+) - g_{y_k,j}(\mathbf{e}_k^+) \geq \delta\Vert\epsilon_+\Vert,
    \end{equation}
    where $\epsilon_+$ is the perturbation caused by similarity loss.
\end{itemize}
By minimizing the objective $L^{\text{sr}}$, the positive center $\overline{\mathbf{e}}_k^+$ shifts its weights toward same-class samples that are more similar to the anchor $\mathbf{e}_k$, causing $\Vert\mathbf{e}_k-\overline{\mathbf{e}}_k^+\Vert$ to decrease $\gamma$. Simultaneously the positive group margin can increase at least $\delta\Vert\epsilon_+\Vert$. Therefore, the classification margin of the $k$-th token can improve at least $m_k' \geq m_k + \delta\Vert\epsilon_+\Vert + \sqrt{2}L_P\gamma$. Therefore, the cross-entropy loss will decrease at least by $\ell_k'\leq\ell_k\cdot\exp\left(-\left(\delta\Vert\epsilon_+\Vert + \sqrt{2}L_P\gamma\right)\right)$, which can also accelerate the pretraining process.

\section{Experiments}
In this section, we show the empirical studies of the proposed \textsc{SimReg} loss. We primarily investigate the advantages in pretraining tasks, including its acceleration on the training loss, improvements of the evaluation on the downstream tasks, and influence the dynamics of the embedding similarity during training. We also examine its sensitivity to hyperparameters and its behavior. Moreover, we explore the practical effects of inserting the \textsc{SimReg} loss at different positions within the model. These experiments can provide useful technical guidance for the community.

\textbf{Model Backbones.} We mainly select LLaMA~\citep{touvron2023llama} and Mixtral~\citep{jiang2024mixtral} as the dense and MoE backbones for pretraining, including the core modules of the mainstream models in the current community, e.g. for RoPE~\citep{su2024roformer}, RMSNorm~\citep{zhang2019root}, and SwiGLU~\citep{shazeer2020glu}. We conduct experiments on dense models with 350M, 1.3B, 3B, and 7B parameters, and on the MoE model with 8B parameters.

\textbf{Training Hyperparameters.} We follow the experimental setups reported in several recent classical LLM pretraining studies~\citep{touvron2023llama,liu2024deepseek,jiang2024mixtral,ernie2025technicalreport} to configure the baseline hyperparameters. We employ the AdamW optimizer~\citep{loshchilov2018decoupled} with $\beta_1=0.9$, $\beta_2=0.95$, and let the weight decay equals to $0.1$. The standard deviation of the weight initialization is set to $0.01$. The global batch size is set to $512$ for the $350$M and MoE-$7$B models, and $2048$ for the $1.3$B, $3$B, and $7$B dense models. The input sequence length is fixed to $2048$. For the learning rate schedule, we adopt a $2000$-step warm-up phase to linearly increase the learning rate from $0$ to $3\times10^{-4}$, followed by a cosine decay strategy that gradually reduces it to one-tenth of its peak value. For dense models, we train about $13$B tokens for the $350$M model and $52$B tokens for the larger dense models. For MoE models, we train approximately $52$B tokens. To avoid loss spikes, we adopt the AdaGC~\citep{wang2025adagc} to clip gradients for all methods. Other details are stated in the Appendix.

\textbf{Baselines.} We select the Simple Contrastive Sentence Embedding~\cite{gao2021simcse}~(SimCSE), Contrastive Pretraining~\cite{neelakantan2022text}~(CPretrain), Consistency Regularization~\cite{yin2023consistency}~(CReg), Similarity Contrastive Estimation~\cite{denize2023similarity}~(SCE). SimCSE adopts the contrastive loss on the sentence embedding. CPretrain minimizes the similarity distribution. CReg treats each token pair as an independent negative example. SCE adopts a weighted similarity via latent distributions. The above works are not all designed for pretraining; however, they share certain conceptual similarities. In our experiments, we uniformly adapt them to the pretraining framework.

\subsection{Empirical Studies on Performance}
\label{exp:emp_studies}

\begin{table*}[t]
\begin{center}
\small
\renewcommand{\arraystretch}{1}
\caption{Generalization performance comparisons: Zero-shot evaluations on the downstream tasks.}
%\small
\setlength{\tabcolsep}{1.5mm}{
\begin{tabular}{@{}l|SSSSSSSSS|c@{}}
\toprule
\multicolumn{1}{c}{} & \multicolumn{1}{c}{{\bf Arc-E}} & \multicolumn{1}{c}{{\bf Arc-C}} & \multicolumn{1}{c}{{\bf BoolQ}} & \multicolumn{1}{c}{{\bf HellaS.}} & \multicolumn{1}{c}{{\bf Obqa}} & \multicolumn{1}{c}{{\bf Piqa}} & \multicolumn{1}{c}{{\bf Mmlu}} & \multicolumn{1}{c}{{\bf WinoG.}} & \multicolumn{1}{c}{{\bf Sciq}} & \multicolumn{1}{c}{{\bf Avg.}} \\
\midrule
\midrule
\textbf{LLaMA-350M}  & 38.64 & 22.95 & 57.09 & 36.51 & 28.40 & 66.49 & 22.95 & 51.30 & 63.20 & 43.06 \\
\midrule
$\circ$ SimCSE & 38.74 & 23.05 & 57.19 & 36.61 & 28.50 & 66.59 & 23.05 & 51.40 & 63.30 & 43.16~(+0.10) \\
$\circ$ CPretrain & 38.20 & 22.80 & 56.60 & 36.10 & 28.10 & 65.90 & 22.80 & 50.80 &  64.26 & 42.84~(-0.22) \\
$\circ$ CReg & 38.90 & 23.20 & 57.40 & 36.80 & 28.70 & 66.90 & 23.20 & 51.80 & 64.15 & 43.45~(+0.39) \\
$\circ$ SCE & 39.10 & 23.40 & 57.80 & 37.10 & 28.90 & 67.20 & 23.40 & 51.80 & 64.24 & 43.66~(+0.60) \\
$\circ$ \textsc{SimReg}     & 40.15 & 24.49 & 57.55 & 37.64 & 29.40 & 68.26 & 22.92 & 52.07 & 64.40 & 44.10~(+1.04) \\
\midrule
\textbf{LLaMA-1.3B}  & 46.21 & 25.09 & 58.01 & 49.60 & 31.80 & 72.14 & 23.07 & 52.80 & 68.90 & 47.51 \\
\midrule
$\circ$ SimCSE & 46.21 & 25.09 & 58.01 & 49.60 & 31.80 & 72.05 & 23.07 & 52.80 & 68.90 & 47.50~(-0.01) \\
$\circ$ CPretrain & 46.16 & 25.04 & 57.96 & 49.55 & 31.64 & 72.09 & 23.02 & 52.75 & 68.73 & 47.43~(-0.08)\\
$\circ$ CReg & 46.41 & 26.29 & 58.31 & 49.90 & 32.10 & 72.64 & 23.27 & 53.10 & 69.00 & 47.78~(+0.27) \\
$\circ$ SCE & 46.71 & 25.59 & 58.41 & 50.10 & 32.10 & 72.54 & 23.27 & 53.10 & 69.64 & 47.94~(+0.43) \\
$\circ$ \textsc{SimReg} & 46.51 & 26.79 & 61.01 & 52.51 & 30.40 & 72.91 & 24.06 & 54.14 & 69.50 & 48.65~(+1.14) \\
\bottomrule
\end{tabular}
}
\label{gen:zero-shot:acc-comparison}
\end{center}
%\vskip -0.15in
\end{table*}

\begin{figure*}[t]
\centering
% ===== 第一行 3 张图 =====
\includegraphics[width=0.328\textwidth]{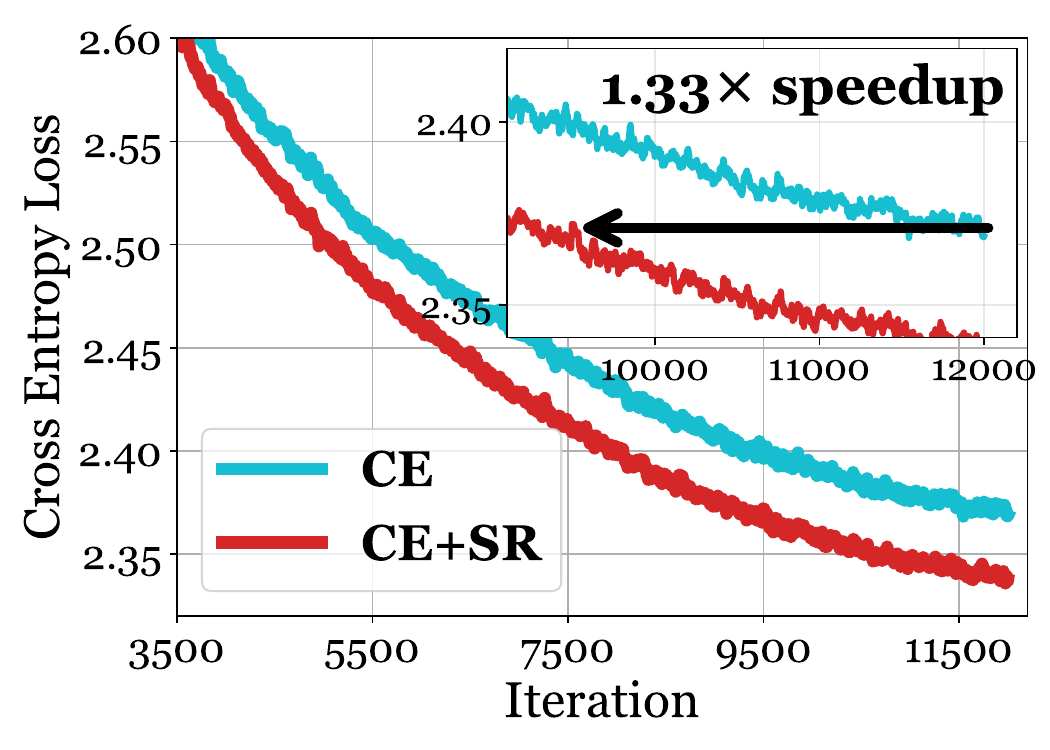}\!\!
\includegraphics[width=0.328\textwidth]{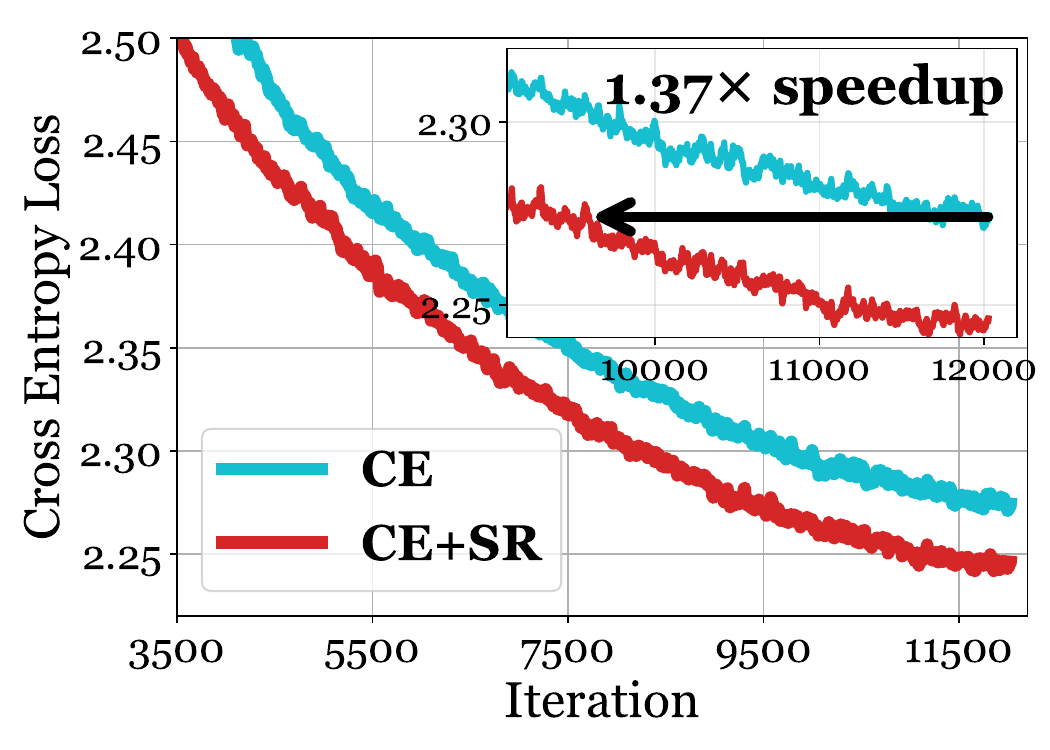}\!\!
\includegraphics[width=0.328\textwidth]{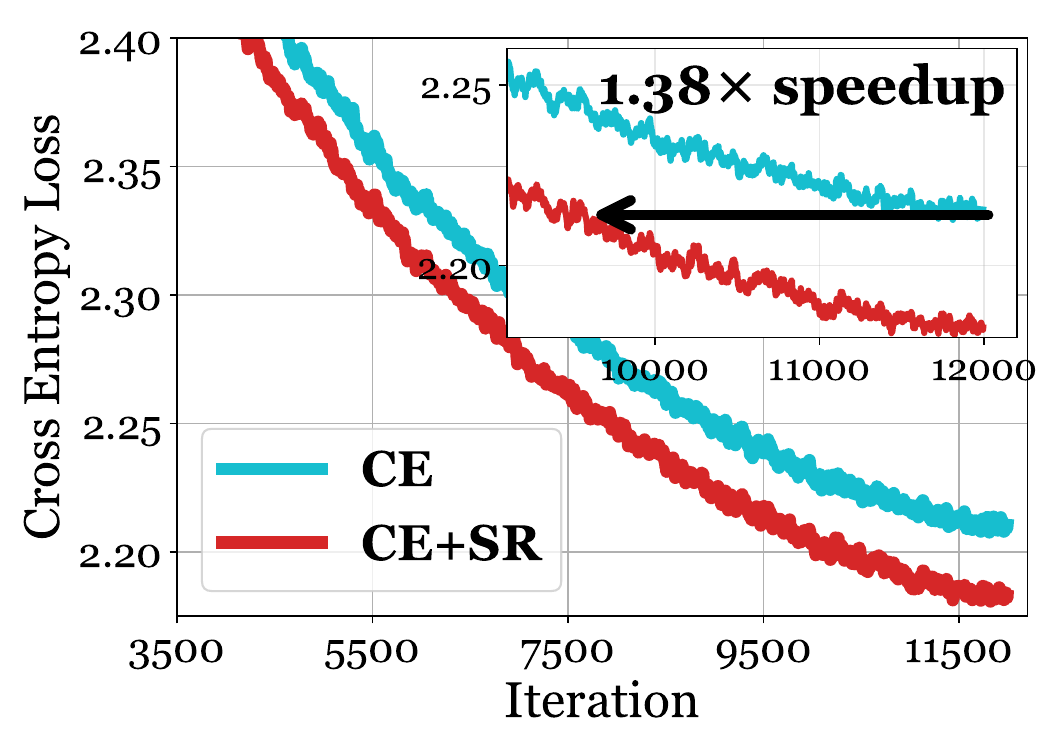}

% 第一行标注 (a)(b)(c)
%\par\small (a)\hspace{0.28\textwidth}(b)\hspace{0.28\textwidth}(c)
%\vspace{0.5em}

% ===== 第二行 3 张图 =====
\includegraphics[width=0.333\textwidth]{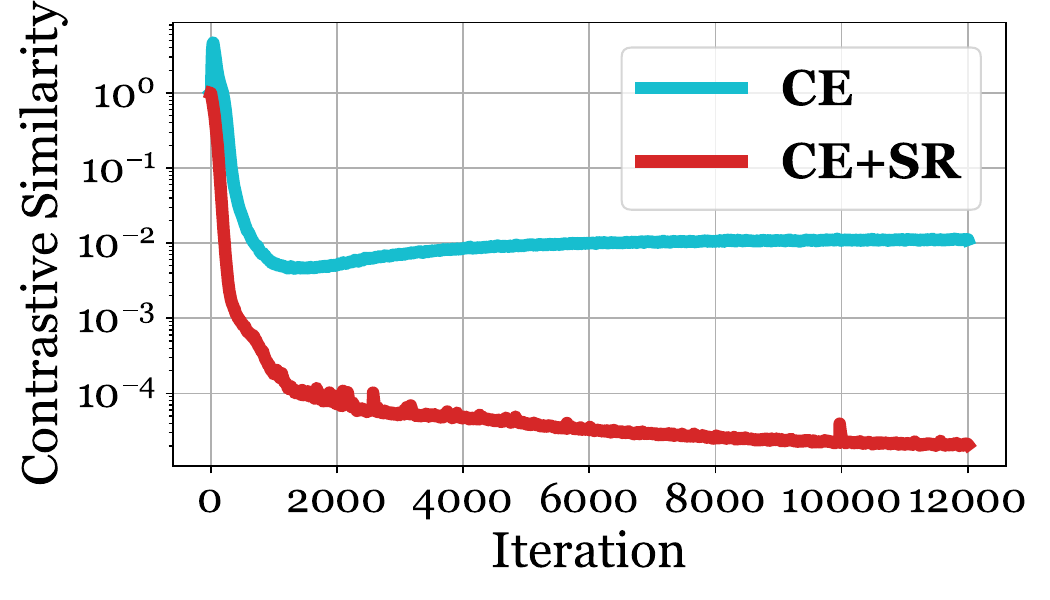}\!\!\!
\includegraphics[width=0.333\textwidth]{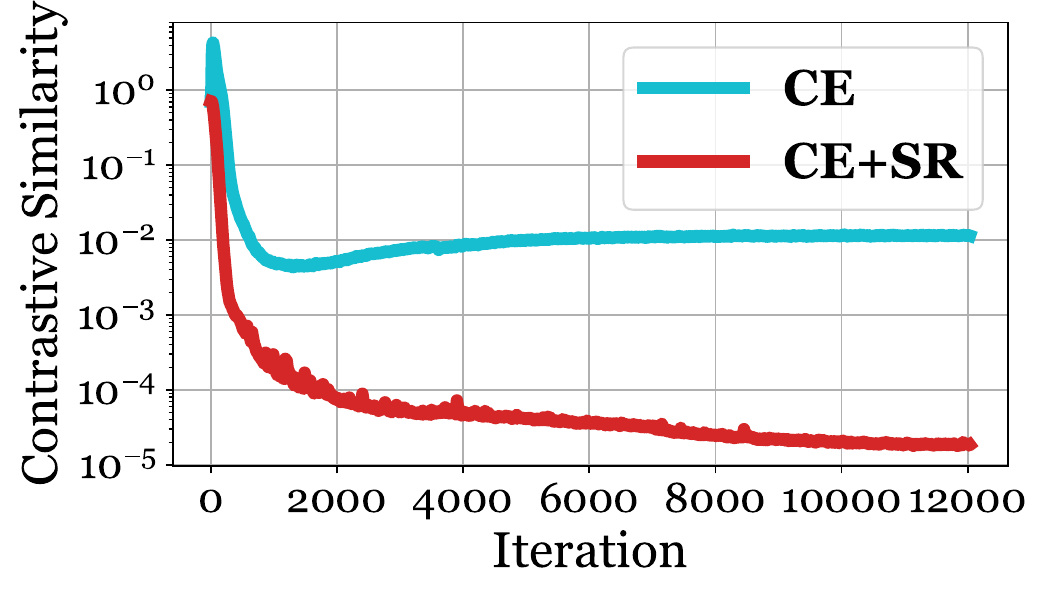}\!\!\!
\includegraphics[width=0.333\textwidth]{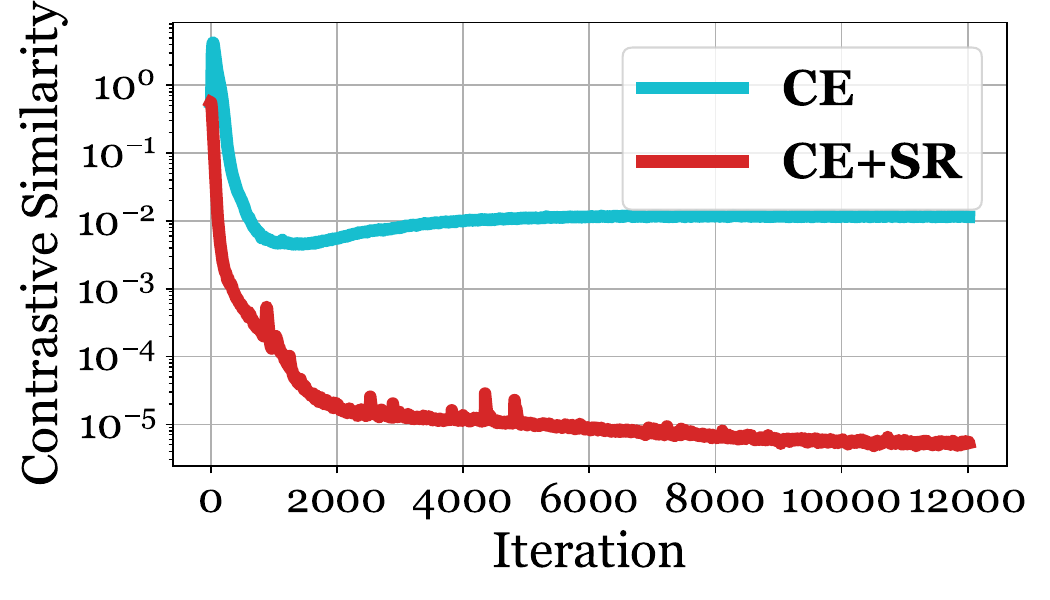}

% 第二行标注 (d)(e)(f)
\par\small 
\makebox[0.32\textwidth][c]{(a) LLaMA-1.3B.}%
\makebox[0.32\textwidth][c]{(b) LLaMA-3B.}%
\makebox[0.32\textwidth][c]{(c) LLaMA-7B.}

\caption{Cross-entropy loss acceleration~(upper) and contrastive similarity improvements~(lower) in the pretraining. ``CE" denotes the cross-entropy loss and ``SR" denotes the similarity regularization loss. \textsc{SimReg} loss helps to further reduce the contrastive similarity.}
\label{fg:opt:ce_loss}
\end{figure*}

\begin{table*}[t]
\begin{center}
\small
\renewcommand{\arraystretch}{1}
\caption{Generalization efficiency: Zero-shot evaluations on the general downstream tasks.}
% \vskip 0.05in
\setlength{\tabcolsep}{1.5mm}{
\begin{tabular}{@{}l|SSSSSSSSS|c@{}}
\toprule
\multicolumn{1}{c}{} & \multicolumn{1}{c}{{\bf Arc-E}} & \multicolumn{1}{c}{{\bf Arc-C}} & \multicolumn{1}{c}{{\bf BoolQ}} & \multicolumn{1}{c}{{\bf HellaS.}} & \multicolumn{1}{c}{{\bf Obqa}} & \multicolumn{1}{c}{{\bf Piqa}} & \multicolumn{1}{c}{{\bf Mmlu}} & \multicolumn{1}{c}{{\bf WinoG.}} & \multicolumn{1}{c}{{\bf Sciq}} & \multicolumn{1}{c}{{\bf Avg.}} \\
\midrule
\midrule
\textbf{LLaMA-350M}  & 38.64 & 22.95 & 57.09 & 36.51 & 28.40 & 66.49 & 22.95 & 51.30 & 63.20 & 43.06 \\
$\circ$ \textsc{SimReg}     & 40.15 & 24.49 & 57.55 & 37.64 & 29.40 & 68.26 & 22.92 & 52.07 & 64.40 & 44.10 \\
$\circ$ \textsc{SimReg}-Chunk  & 39.77 & 24.23 & 58.14 & 37.25 & 29.40 & 67.59 & 23.02 & 51.84 & 64.40 & 43.96 \\
\midrule
\textbf{LLaMA-1.3B}  & 46.21 & 25.09 & 58.01 & 49.60 & 31.80 & 72.14 & 23.07 & 52.80 & 68.90 & 47.51 \\
$\circ$ \textsc{SimReg}     & 46.51 & 26.79 & 61.01 & 52.51 & 30.40 & 72.91 & 24.06 & 54.14 & 69.50 & 48.65\\
$\circ$ \textsc{SimReg}-Chunk & 46.80 & 26.11 & 59.17 & 51.94 & 31.80 & 72.25 & 23.12 & 54.78 & 69.00 & 48.33 \\
\midrule
\textbf{LLaMA-3B}    & 48.91 & 27.30 & 58.29 & 55.67 & 33.00 & 74.16 & 23.65 & 55.49 & 73.50 & 50.00 \\
$\circ$ \textsc{SimReg}     & 50.59 & 28.07 & 58.65 & 57.65 & 33.40 & 74.32 & 23.95 & 56.67 & 75.30 & 50.96\\
$\circ$ \textsc{SimReg}-Chunk  & 50.80 & 27.39 & 62.48 & 58.49 & 33.60 & 73.88 & 22.95 & 55.64 & 73.20 & 50.94  \\
\midrule
\textbf{LLaMA-7B}    & 53.07 & 28.84 & 54.07 & 60.41 & 33.80 & 76.12 & 23.79 & 57.30 & 75.70 & 51.45 \\
$\circ$ \textsc{SimReg}     & 52.57 & 29.01 & 59.79 & 62.01 & 35.80 & 75.14 & 24.47 & 59.04 & 76.20 & 52.67 \\
$\circ$ \textsc{SimReg}-Chunk & 51.60 & 29.69 & 62.39 & 61.80 & 35.80 & 75.46 & 23.51 & 58.72 & 76.00 & 52.77 \\
\midrule
\midrule
\textbf{Mixtral-8$\times$1B} & 48.86 & 29.18 & 54.62 & 59.57 & 34.00 & 73.88 & 24.17 & 56.99 & 72.40 & 50.41 \\
$\circ$ \textsc{SimReg}             & 51.81 & 28.75 & 60.03 & 62.53 & 35.00 & 75.08 & 23.59 & 54.30 & 74.10 & 51.69 \\
$\circ$ \textsc{SimReg}-Chunk       & 52.04 & 28.98 & 60.26 & 62.76 & 35.23 & 73.88 & 23.82 & 54.53 & 73.10 & 51.62 \\
\bottomrule
\end{tabular}
}
\label{gen:zero-shot:acc}
\end{center}
%\vskip -0.15in
\end{table*}

\textbf{Higher Generalization.} Table~\ref{gen:zero-shot:acc-comparison} shows the zero-shot generalization results on a range of downstream tasks. Overall, existing consistency-based baselines bring only marginal or inconsistent improvements in the averaged performance across tasks, and in some cases even lead to slight degradation. In contrast, our method achieves the most consistent and significant gains in terms of average accuracy for both model scales, improving the mean score by +1.04\% for LLaMA-350M and +1.14\% for LLaMA-1.3B. This trend indicates that our approach provides more effective downstream transfer and stronger generalization performance than prior methods under the general downstream tasks. 

\textbf{Higher Convergence.} We first demonstrate the training acceleration of \textsc{SimReg} in large-scale pretraining tasks. As shown in Figur~\ref{fg:opt:ce_loss} upper part, on the 1.3B model, the speedup can reach nearly 33\%, and after training on 52B tokens, the cross entropy loss can be reduced by about 0.05. On larger-scale models, including the 3B model and the 7B model, \textsc{SimReg} achieves more than 37\% speedup when training reaches 52B tokens, with the final training loss reduced by about 0.03. In the lower part, we present the \textsc{SimReg} loss. It can be observed that cross-entropy does not impose a mandatory constraint on feature similarity. When training with cross-entropy alone, the feature similarity undergoes a rapid decline in the early stage, and then gradually tends to stabilize. At this point, the network no longer additionally learns to accelerate classification training by enhancing feature separability. An interesting phenomenon we observe is that, when trained solely with cross-entropy, the similarity regularization value for almost all networks eventually converges to around 0.01, which implies that the average angle between words of different classes is approximately 61.3 degrees. After introducing the \textsc{SimReg} loss, the embedding similarity decreases significantly, with the regularization loss converging to about 0.00001, indicating that the average angle achieves approximately 74 degrees among tokens.

% \begin{table}[b]
% \small
%     %\vspace{-0.75cm}
%     \caption{Performance of different chunk size.}
%     %\vspace{0.1cm}
%     \label{tb:chunk}
%     \centering
%     \begin{tabular}{c|cccccc}
%         \toprule
%               & 128 & 256 & 512 & 1024 & 2048 \\
%         \midrule
%         Loss  & 2.194 & 2.189 & 2.186 & \textbf{2.184} & 2.186 \\
%         \midrule
%         Avg. Acc.  & 51.46 & 52.52 & 52.54 & \textbf{52.77} & 52.67 \\
%         \bottomrule
%     \end{tabular}
%     %\vspace{-0.4cm}
% \end{table}

\textbf{Chunk-wise \textsc{SimReg} v.s. Full \textsc{SimReg}.} In Table~\ref{gen:zero-shot:acc}, it can be observed that the chunk-wised \textsc{SimReg} achieves comparable performance to that of Full-\textsc{SimReg}, and even outperforms it on the 7B model. Under large chunks, the expressive capacity of the \textsc{SimReg} loss becomes limited. When dealing with an excessively large number of tokens, the effective supervisory signal for each individual token is weakened. There exists a trade-off between the expressive capacity of the loss and its strength of supervision. This phenomenon becomes more pronounced as the parameter scale increases. As the model scales up, the dimensionality of the hidden states grows proportionally, which naturally leads to larger angles between embeddings. When computing similarity regularization in high-dimensional spaces, the number of participating tokens has a stronger influence on the evaluation quality for each individual token. Thus, chunk-wise \textsc{SimReg} can be considered as an effective alternative to full \textsc{SimReg} for the large-scale model pretraining.

\textbf{Scaling to Large Models.} Table~\ref{gen:zero-shot:acc} reports the performance on larger-sacle models. Overall, introducing the \textsc{SimReg} loss consistently improves the average performance from 350M to 7B. \textsc{SimReg} can bring +1.14\% average improvement on LLaMA-1.3B, +0.96\% on LLaMA-3B, +1.22\% on LLaMA-7B, and +1.28\% on Mixtral-8$\times$1B. These results highlight that \textsc{SimReg} provides stable and non-trivial gains as the model scale increases. Moreover, \textsc{SimReg} achieves the largest single-task gain of +5.72\% on BoolQ with LLaMA-7B. Besides BoolQ, we also observe clear improvements on HellaSwag, WinoGrande, and SciQ across multiple scales, showing that it is particularly effective for reasoning-heavy and multi-choice tasks. These consistent improvements further suggest that \textsc{SimReg} is a simple yet broadly applicable strategy for the large-scale pretraining to enhance generalization.

\subsection{Hyperparameter Sensitivity}
\label{exp:hyperparameter sensitivity}

\begin{figure*}[t]
\centering
    \subfloat[Grid Search of $\left(\tau, \lambda\right)$.]{
	\includegraphics[width=0.285\textwidth]{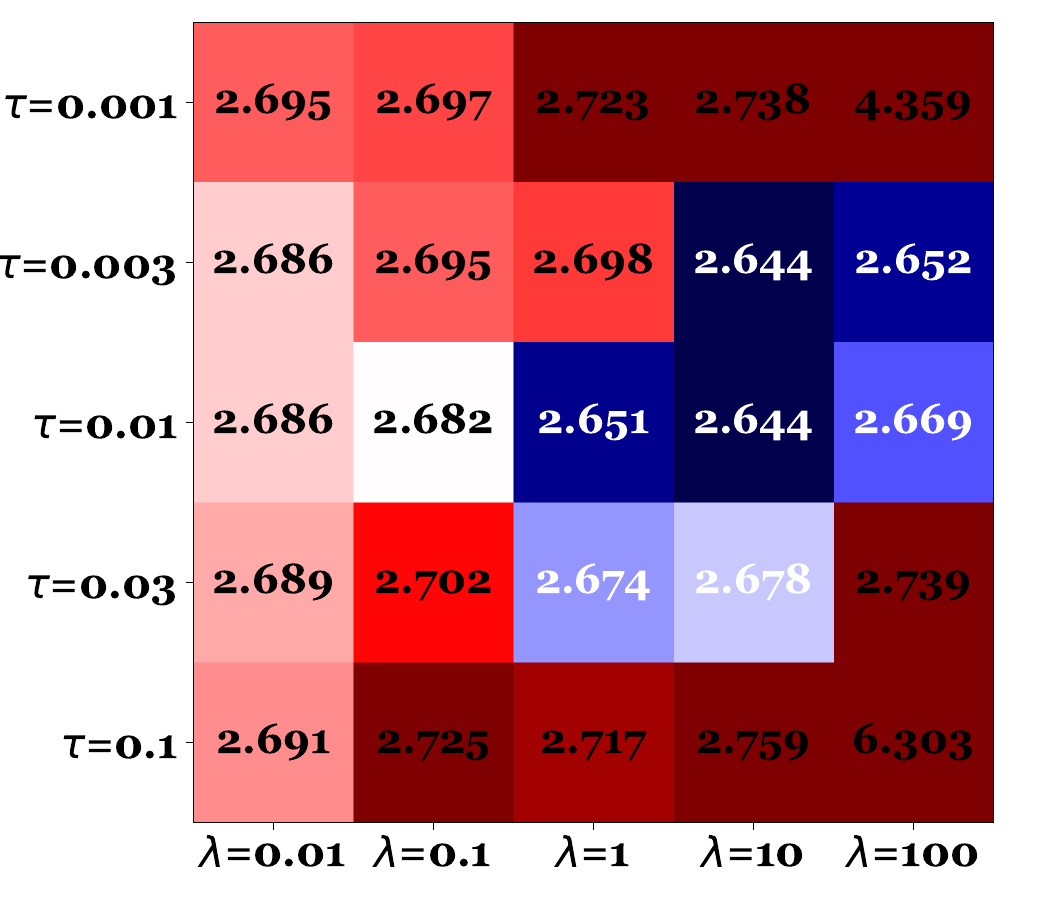}}
    \subfloat[Fine-grained Search on $\lambda$.]{
	\includegraphics[width=0.32\textwidth]{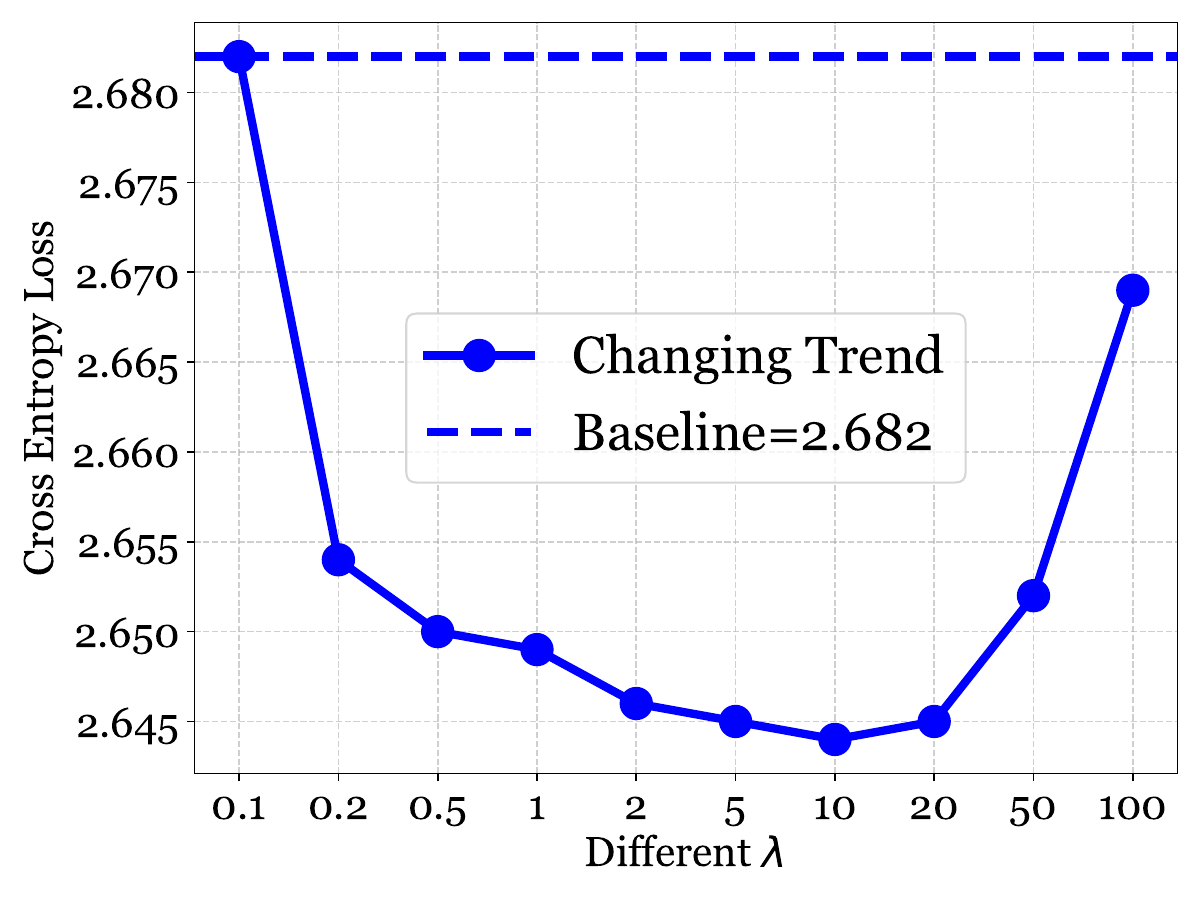}}
    \subfloat[Scaling with Model Size.]{
	\includegraphics[width=0.322\textwidth]{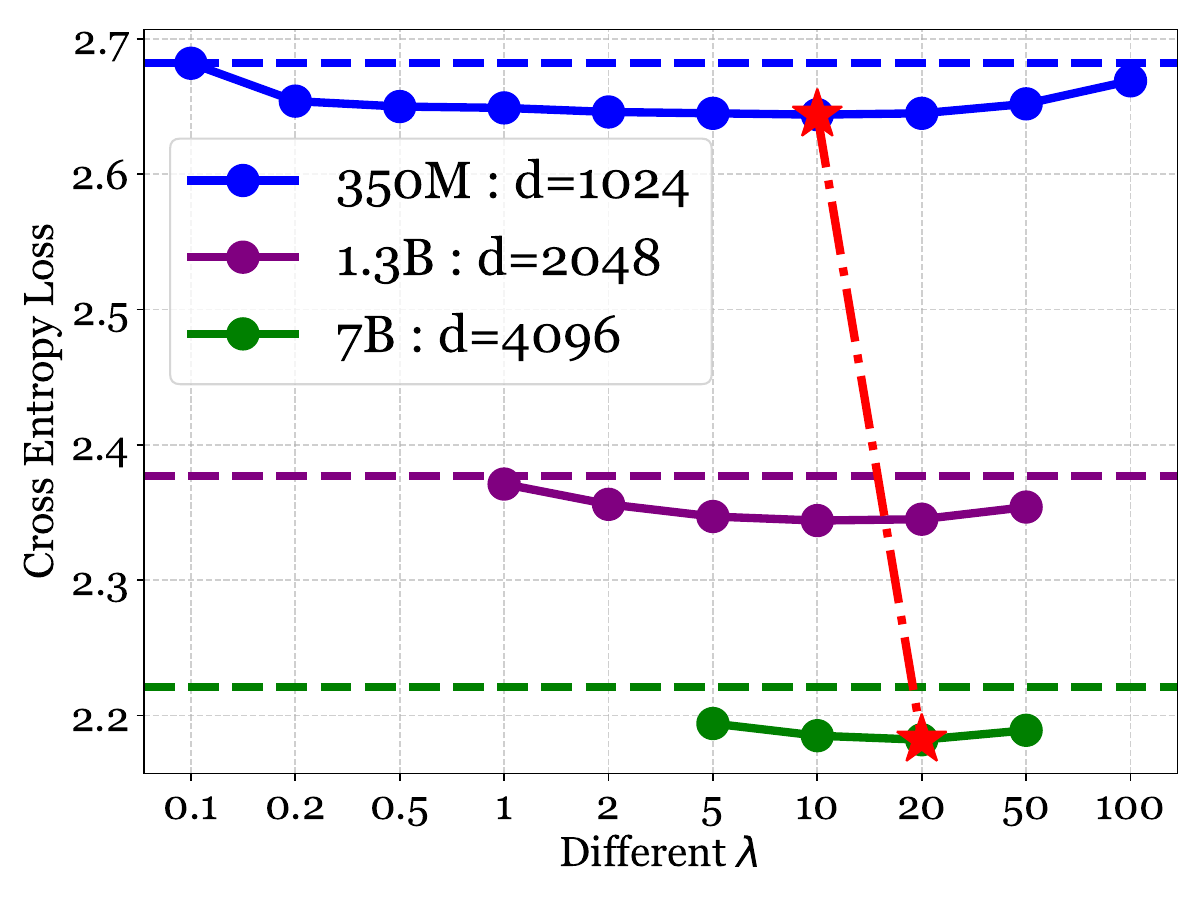}}
%\vskip -0.1in
\caption{(a) Grid search over hyperparameters $\tau$ and $\lambda$. The blue blocks indicate the values where the final training loss under the corresponding combination~$\left(\tau,\lambda\right)$ is lower than baseline, with darker colors representing lower losses. (b) We further conduct a fine-grained search over different $\lambda$ values at the generally optimal $\tau=0.01$, using an approximate $2\times$ scaling ratio. (c) We explore the trends on different $\lambda$ across different model sizes~(the red line indicates the optimal trend).}
\label{fg:hyperparameters sensitivity}
\end{figure*}
We first grid search $(\tau,\lambda)$ on the 350M model to identify a valid range, followed by a fine-grained search to determine their optimal combinations. Subsequently, we conduct scaling experiments on the 1.3B and 7B models to examine how the optimal choices vary as the model size increases and the corresponding token embedding dimension grows. As shown in Figure~\ref{fg:hyperparameters sensitivity}~(a), to explore the stable results, we grid search the temperature coefficient $\tau$ from $\left[0.001, 0.003, 0.01, 0.03, 0.1\right]$ with a $3\times$ skip, and coarsely choose the coefficient $\lambda$ from $\left[0.01, 0.1, 1, 10, 100\right]$ with a $10\times$ skip. The valid range for $\tau$ is relatively limited, with $0.01$ proving to be a robust selection for all models. $\lambda$ spans a broad effective range from $0.1$ to $100$. Figure~\ref{fg:hyperparameters sensitivity}~(b) presents a fine-grained exploration of $\lambda$, varying it from $0.1$ to $100$ with roughly $2\times$ resolution. The results reveal a stable region between $2$ and $20$. In Figure~\ref{fg:hyperparameters sensitivity}~(c), we explore the scaling of hyperparameters and infer from results across different model sizes how to select optimal hyperparameters. Specifically, when the embedding dimension increases, each token is represented in a higher-dimensional space. Therefore, it becomes necessary to increase $\lambda$ to maintain training efficiency. Our experiments confirm this trend, and current results suggest that every time the embedding dimension doubles, the optimal hyperparameter increases by approximately a factor of $\sqrt{2}$. The optimal $\tau$ can be fixed as $0.01$ for all models.

\subsection{Optimal Position of Adopting \textsc{SimReg}}
\label{exp:position}

\begin{wrapfigure}[12]{r}{0.5\textwidth}
\centering
\vspace{-0.5cm}
    \includegraphics[width=0.48\textwidth]{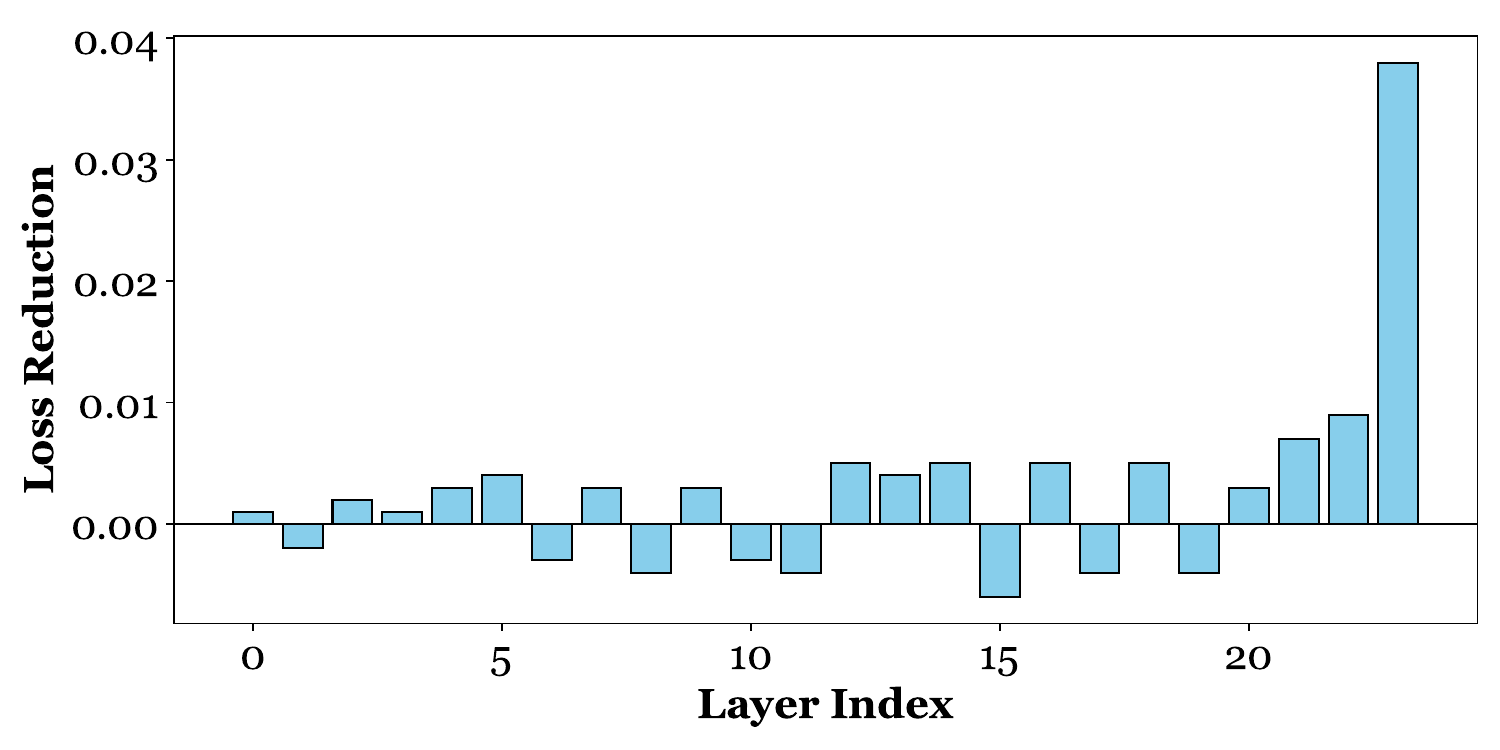}
\caption{Loss changes of adopting our \textsc{SimReg} loss at different layers on 1B model.}
\label{fg:optimal position}
\end{wrapfigure}

In this part, we empirically investigate at which positions in the model embedding supervision yields the best results. We divide the network according to its natural layer-wise structure and apply supervision at different depths. As shown in Figure~\ref{fg:optimal position}, supervision on intermediate layers brings only negligible performance gains. This is expected, as token representations in the middle layers are not simply tied to independent word meanings; instead, they encode blended contextual information aggregated from preceding tokens. Enforcing similarity regularization on such entangled representations may therefore provide limited useful signal. In contrast, the final layers gradually project these broad contextual representations into more distinct semantic spaces that are directly used for next-token prediction. Our experiments further show that applying \textsc{SimReg} only at the last layer is sufficient to achieve efficient pretraining.

\subsection{Runtime and Memory Consumptions}
\label{exp:training overhead}

\begin{wraptable}[8]{r}{0.6\textwidth}
    \vspace{-0.4cm}
    \centering
    \caption{Performance and sensitivity ($T=600$).}
    \label{tb:overhead}
    \begin{tabular}{c|cc}
        \toprule
              & Time~(s / iter) & Memory~(GB) \\
        \midrule
        w/o \textsc{SimReg}      & 2.295 & 373.45 \\
        w/ \textsc{SimReg}       & 2.296 & 375.28 \\
        w/ \textsc{SimReg}-Chunk & 2.295 & 373.84 \\
        \bottomrule
    \end{tabular}
\end{wraptable}

We evaluate the training efficiency of our method on a 7B-scale model with a token embedding dimension of 4096. For \textsc{SimReg}-Chunk, we use a chunk size of 1024 to further reduce the computational footprint. All reported statistics are collected on H800 GPUs, and memory usage is measured by the maximum GPU memory allocation. As shown in Table~\ref{tb:overhead}, incorporating the full \textsc{SimReg} loss results in less than a 2\% increase in runtime and under a 1\% increase in GPU memory consumption. In comparison, \textsc{SimReg}-Chunk introduces only negligible computational and memory overhead, making it effectively resource-neutral in practice. These results show that \textsc{SimReg} delivers meaningful performance gains with minimal additional training cost, highlighting its practicality as a lightweight and effective auxiliary component for large-scale pretraining.

\section{Conclusion}

In this work, we introduced \textsc{SimReg}, a similarity regularization loss for large-scale pretraining. We show that cross-entropy alone does not sufficiently enforce embedding consistency, whereas \textsc{SimReg} strengthens representation learning by aligning same-class tokens while separating different classes. Experiments on both dense and MoE models demonstrate that \textsc{SimReg} consistently accelerates convergence by more than 30\% and improves downstream performance by over 1\%. Moreover, it remains robust across different model scales and hyperparameter settings, indicating its practical applicability. These findings highlight consistency regularization as a promising direction for improving the efficiency and generalization of LLM pretraining.

%%%%%%%%%%%%%%%%%%%%%%%%%%%%%%%%%%%%%%%%%%%%%%%%%%%%%%%%%%%%
% Reference
\newpage

% \textbf{Statement of Broader Impacts and Limitations}

% This work aims to improve the pre-training efficiency of large models, with a primary focus on theoretical frameworks such as representation learning and contrastive learning. It does not introduce deployed systems or applications beyond this scope, and therefore we do not anticipate additional societal impacts. As such, it does not directly introduce new deployed systems or applications, and we do not anticipate immediate negative societal impacts. The potential positive impact is to provide guidance for designing more computation- and memory-efficient optimization algorithms, especially in settings where full-parameter updates are expensive.

{
\small
\bibliographystyle{plainnat}
\bibliography{reference}
}
%%%%%%%%%%%%%%%%%%%%%%%%%%%%%%%%%%%%%%%%%%%%%%%%%%%%%%%%%%%%

%%%%%%%%%%%%%%%%%%%%%%%%%%%%%%%%%%%%%%%%%%%%%%%%%%%%%%%%%%%%

%%%%%%%%%%%%%%%%%%%%%%%%%%%%%%%%%%%%%%%%%%%%%%%%%%%%%%%%%%%%%%%%%%%%%%%%%%%%%%%
%%%%%%%%%%%%%%%%%%%%%%%%%%%%%%%%%%%%%%%%%%%%%%%%%%%%%%%%%%%%%%%%%%%%%%%%%%%%%%%
% APPENDIX
%%%%%%%%%%%%%%%%%%%%%%%%%%%%%%%%%%%%%%%%%%%%%%%%%%%%%%%%%%%%%%%%%%%%%%%%%%%%%%%
%%%%%%%%%%%%%%%%%%%%%%%%%%%%%%%%%%%%%%%%%%%%%%%%%%%%%%%%%%%%%%%%%%%%%%%%%%%%%%%
\newpage
\appendix
\onecolumn

\section{Appendix: Experiments}
\subsection{Experimental Setups}
\label{ap:exp setups}
Here we present the detailed experimental setups in this paper to ensure the reproducibility.

\textbf{Model Hyperparameters.} We mainly select LLaMA2~\citep{touvron2023llama} and Mixtral~\citep{jiang2024mixtral} as the dense and MoE backbones for pretraining, including the core modules of the mainstream models in the current community, e.g. for RoPE~\citep{su2024roformer}, RMSNorm~\citep{zhang2019root}, and SwiGLU~\citep{shazeer2020glu}. We follow the common practices in the community to scale models of different sizes, and the detailed configurations are shown in Table~\ref{ap:tab:model config}.
\begin{table}[H]
\centering
\caption{Model Hyperparameters.}
\begin{tabular}{c|ccccc}
\toprule
 & Experts & Layers & Attention heads & Embedding dim & FFN hidden size \\
\midrule
 LLaMA$2$-$350$M          & 1 & 24 & 16 & 1024 &  2371 \\
 LLaMA$2$-$1.3$B          & 1 & 24 & 32 & 2048 &  5461 \\
 LLaMA$2$-$3$B            & 1 & 26 & 32 & 3072 &  8640 \\
 LLaMA$2$-$7$B            & 1 & 32 & 32 & 4096 & 11008 \\
\midrule
 Mixtral-8$\times$1B    & 8 & 24 & 32 & 2048 &  5632 \\
\bottomrule
\end{tabular}
\label{ap:tab:model config}
\end{table}

\textbf{Training Hyperparameters.} We follow the experimental setups reported in several recent classical LLM pretraining studies~\citep{touvron2023llama, liu2024deepseek, jiang2024mixtral} to configure the baseline hyperparameters, ensuring comparability with prior work. Specifically, we employ the AdamW optimizer~\citep{loshchilov2018decoupled} with $\beta_1=0.9$ and $\beta_2=0.95$, and a weight decay of $0.1$. The standard deviation of weight initialization is set to $0.01$. To balance efficiency and stability, we use a global batch size of 
$512$ for the 350M and MoE-1$\times$8B models, and $2048$ for the 1.3B, 3B, and 7B dense models, while the input sequence length is fixed at $2048$.

For the learning rate schedule, we adopt a 2000-step warm-up phase that linearly increases the learning rate from $0$ to $3e\text{-}4$, followed by a cosine decay strategy that gradually reduces it to one-tenth of its peak value. Regarding training length, dense models are trained for $12,500$ steps, corresponding to roughly $13$B tokens for the $350$M model and $52$B tokens for the larger dense models. In contrast, MoE models are trained for $50,000$ steps to ensure comparable exposure of approximately $52$B tokens. Finally, to mitigate potential instabilities caused by loss spikes, we adopt AdaGC~\citep{wang2025adagc} for adaptive gradient clipping. We summarize the details in Table~\ref{tab:hyperparams}.
\begin{table}[H]
\centering
\caption{Training Hyperparameters.}
\begin{tabular}{c|ccccccccc}
\toprule
 & batchsize & seqlen & learning rate & $\lambda_{\text{w}}$ & $\beta_1$ & $\beta_2$ & clip-$\lambda$ & clip-$\beta$\\
\midrule
 LLaMA-350M   & 512  & 2048 & 4e-4$\ \rightarrow \ $4e-5 & 0.1 & 0.9 & 0.95 & 1.04 & 0.99 \\
 LLaMA-1.3B   & 2048 & 2048 & 3e-4$\ \rightarrow \ $3e-5 & 0.1 & 0.9 & 0.95 & 1.04 & 0.99 \\
 LLaMA-3B     & 2048 & 2048 & 3e-4$\ \rightarrow \ $3e-5 & 0.1 & 0.9 & 0.95 & 1.04 & 0.99 \\
 LLaMA-7B     & 2048 & 2048 & 3e-4$\ \rightarrow \ $3e-5 & 0.1 & 0.9 & 0.95 & 1.04 & 0.99 \\
 Mixtral-8$\times$1B & 512  & 2048 & 3e-4$\ \rightarrow \ $3e-5 & 0.1 & 0.9 & 0.95 & 1.04 & 0.99 \\
\bottomrule
\end{tabular}
\label{tab:hyperparams}
\end{table}

\textbf{Specific Hyperparameters.} Our proposed loss function is primarily characterized by two key hyperparameters, the temperature $\tau$ and the coefficient $\lambda$. We conduct extensive grid search experiments~($\tau\in\left[0.001, 0.01, 0.1\right]$ and $\lambda\in\left[0.2, 0.5, 1, 2, 5, 10, 20, 50, 100\right]$) on the $350$M model to determine the effective range of these hyperparameters, and validate them on larger models according to scaling theory. The simple settings of $\tau=0.01$ and $\lambda=10$ are sufficient to achieve good performance for most experiments. To better adapt to the model scaling, we explore a more refined yet simple strategy to determine the selections, which is detailed in Sec.\ref{ap:scale hyperparameters}.

\textbf{Evaluations.} To ensure a fair comparison, we conduct all evaluations on EleutherAI/lm-evaluation benchmark~\citep{eval-harness}. We mainly evaluate the performance of the pretrained model on downstream tasks of arc\_easy, arc\_challenge, openbookqa, boolq, hellaswag, piqa, winogrande, mmlu, sciq~(general reasoning ability) and the domain-specific downstream tasks of gsm8k, drop, race, squadv2, nq\_open, humaneval, mbpp~(three domains: math, code, and reading comprehension).

\textbf{Training Resources.} We conduct experiments on H$800$ GPUs. Pretraining the $350$M model on $13$B tokens requires approximately $56$ GPU hours per experiment and the $7$B model on $52$B tokens takes over $2{,}000$ GPU hours per experiment.

\subsection{How to Scale Hyperparameters on Large Models?}
\label{ap:scale hyperparameters}
In this part, we introduce a refined hyperparameter tuning mechanism to accommodate model scaling. Before introducing it, we first demonstrate the relationship between the representation ability of our \textsc{SimReg} loss and the dimensionality of embeddings in the model. The \textsc{SimReg} loss regularizes pretraining by leveraging the token embedding similarity between pairs of tokens. By assuming $\mathbf{x}, \mathbf{y} \in \mathbb{R}^{d}$ are independent and identically distributed as isotropic random variables, e.g., $\mathbf{x}, \mathbf{y} \sim \mathcal{N}\left(0,I_d\right)$. Thus, we consider their cosine similarity $z=\frac{\left\langle\mathbf{x},\mathbf{y}\right\rangle}{\Vert\mathbf{x}\Vert\cdot\Vert\mathbf{y}\Vert}\in\left[-1,1\right]$. Without loss of generality, we can assume $\frac{\mathbf{y}}{\Vert\mathbf{y}\Vert}=\left(1,0,\cdots,0\right)$ as the first basis of the spherical space $S^{d-1}$. Then the distribution of $z$ can be transferred to the study of the first coordinate of $v\sim \text{Uinf}\left(S^{d-1}\right)$. Substitute $v$ into the iterative form of spherical coordinates $v=\left(\cos{\theta}, \sin{\theta}\cdot \zeta\right)$ where $\zeta\in S^{d-2}$. According to the decomposition of the spherical surface unit, we have $d\sigma_{d-1}(v)=\sin^{d-2}(\theta) \ d\theta \ d\sigma_{d-1}(\zeta)$ and the marginal density of the polar angle:
\begin{align*}
    f_p(\theta)=\frac{1}{\vert S^{d-1}\vert}\int_{S^{d-2}}\sin^{d-2}(\theta) \ d\sigma_{d-2}(v)= \frac{\vert S^{d-2}\vert}{\vert S^{d-1}\vert}\sin^{d-2}(\theta).
\end{align*}
Then we consider the variable $z$. Due to the first coordinate $z=v_0=\cos(\theta)$, we have:
\begin{align*}
    f_p(z)
    =f_p(\theta)\left\vert\frac{d\theta}{dz}\right\vert=\frac{\vert S^{d-2}\vert}{\vert S^{d-1}\vert}\cdot\frac{\sin^{d-2}(\theta)}{\sin(\theta)}
    =\frac{\vert S^{d-2}\vert}{\vert S^{d-1}\vert}\left(1-z^2\right)^\frac{d-3}{2}
    =\frac{\Gamma(\frac{d}{2})}{\sqrt{\pi}\Gamma(\frac{d-1}{2})}\left(1-z^2\right)^\frac{d-3}{2}.
\end{align*}
It is easy to check that $\mathbb{E}\left[z\right]=0$ and $\mathbb{E}\left[z^2\right]=\frac{1}{d}$. Therefore, as the model size increases and the embedding dimensionality changes from $d_0$ to $d_1$, the capacity of \textsc{SimReg} loss decreases by a factor of $\sqrt{\frac{d_1}{d_0}}$. To preserve the representation capability, we can revise the $\lambda$ coefficient.

We next investigate the feasibility of this scaling method from an empirical perspective. We separately sweep the hyperparameters and report the evaluation perplexity (ppl) at the end of training.
\begin{table}[H]
\centering
\caption{Validation perplexity~(generalization performance) of different $(\tau,\lambda_\text{reg})$.}
\begin{tabular}{c|ccccc}
\toprule
 & $\tau=0.005$ & $\tau=0.01$ & $\tau=0.02$ & $\tau=0.05$ & $\tau=0.1$ \\
\midrule
 $\lambda_{\text{reg}}=0$ (baseline) & \multicolumn{5}{c}{15.06} \\
\midrule
 $\lambda_{\text{reg}}=0.1$ & 14.98 & 15.01 & 14.92 & 14.95 & 14.98 \\
 $\lambda_{\text{reg}}=0.2$ & 14.50 & 14.38 & 14.36 & 14.79 & 14.81 \\
 $\lambda_{\text{reg}}=0.5$ & 14.45 & 14.41 & 14.43 & 14.54 & 14.88 \\
 $\lambda_{\text{reg}}=1$   & 14.41 & 14.45 & 14.46 & 14.67 & 14.71 \\
 $\lambda_{\text{reg}}=2$   & 14.39 & 14.41 & 14.49 & 14.65 & 14.74 \\
 $\lambda_{\text{reg}}=5$   & 14.36 & 14.36 & 14.37 & 14.35 & 15.13 \\
 $\lambda_{\text{reg}}=10$  & 14.34 & \underline{\textbf{14.25}} & 14.36 & 14.44 & 15.65 \\
 $\lambda_{\text{reg}}=20$  & 14.41 & 14.29 & 14.44 & 14.59 & - \\
 $\lambda_{\text{reg}}=50$  & 14.74 & 14.53 & 14.55 & 14.78 & - \\
\bottomrule
\end{tabular}
\label{tab:llama-350m}
\end{table}
The optimal range and variation trend in Table~\ref{tab:llama-350m} and Figure~\ref{fg:hyperparameters sensitivity} are almost identical to those observed in the optimization process, \textbf{indicating that the improvements brought by the \textsc{SimReg} loss in both optimization and generalization are consistent}. The optimal choice of $\tau$ remains concentrated around $0.01$. Next, we evaluate models of different scales (primarily with increased embedding hidden sizes), while keeping $\lambda$ fixed at $0.01$.
\begin{table}[H]
\centering
\caption{Optimal validation perplexity~(generalization performance) of different model size.}
\begin{tabular}{c|cccc}
\toprule
 & 350M~($d=1024$) & 1.3B~($d=2048$) & 3B~($d=3072$) & 7B~($d=4096$)  \\
\midrule
 $\lambda_{\text{reg}}=0$ (baseline) & 15.06 & 10.72 & 9.70 & 8.99  \\
\midrule
 $\lambda_{\text{reg}}=5$   & 14.36 & 10.46 & 9.50 & 8.92 \\
 $\lambda_{\text{reg}}=10$  & \underline{\textbf{14.25}} & \underline{\textbf{10.41}} & 9.46 & 8.84 \\
 $\lambda_{\text{reg}}=20$  & 14.29 & 10.42 & \underline{\textbf{9.44}} & \underline{\textbf{8.78}} \\
 $\lambda_{\text{reg}}=50$  & 14.33 & 10.49 & 9.49 & 8.81 \\
\bottomrule
\end{tabular}
\label{tab:change_trend}
\end{table}
It can be observed that the trend largely aligns with our hypothesis. Therefore, we propose the following estimation method for the optimal hyperparameters:
\begin{align*}
    \tau = 0.01, \ \lambda_{\text{reg}}\approx 10\times\sqrt{\frac{d}{1024}},
\end{align*}
where $d$ is the dimension of the hidden-size of the token embedding. Of course, the scale of the model also affects the results. In practice, a simple grid search within this range of choices can be performed to identify the optimal combination.

\newpage
\subsection{SimReg Loss Curves}
In this section, we mainly present the variations of the \textsc{SimReg} loss. We explore the limitations of cross-entropy in LLM pretraining, namely, that it cannot achieve better classification performance simply by further reducing feature separability. This is because cross-entropy focuses solely on aligning predictions with ground-truth labels, while leaving the underlying structure of token embeddings insufficiently constrained. As the model scales up, this weakness becomes more pronounced: embeddings of the same class may still scatter in the representation space, leading to instability in optimization and slower convergence. By contrast, the \textsc{SimReg} loss explicitly regularizes intra-class consistency and inter-class separation, complementing cross-entropy with a more direct control of embedding geometry. This additional constraint not only improves convergence speed but also yields more robust generalization in downstream tasks.

\begin{figure}[H]
\centering
    \subfloat[LLaMA-350M.]{
	\includegraphics[width=0.45\textwidth]{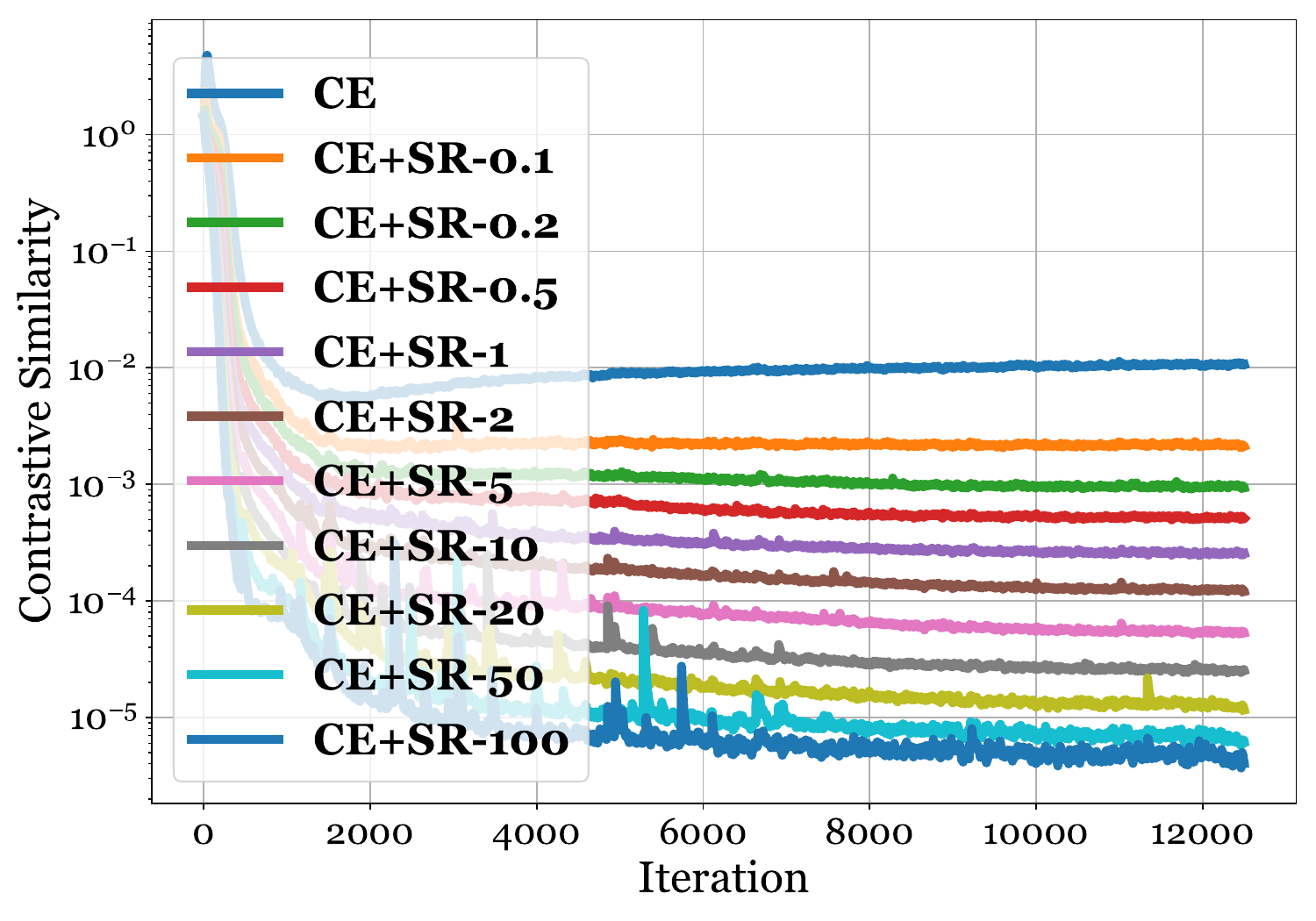}}
    \subfloat[LLaMA-1.3B.]{
	\includegraphics[width=0.45\textwidth]{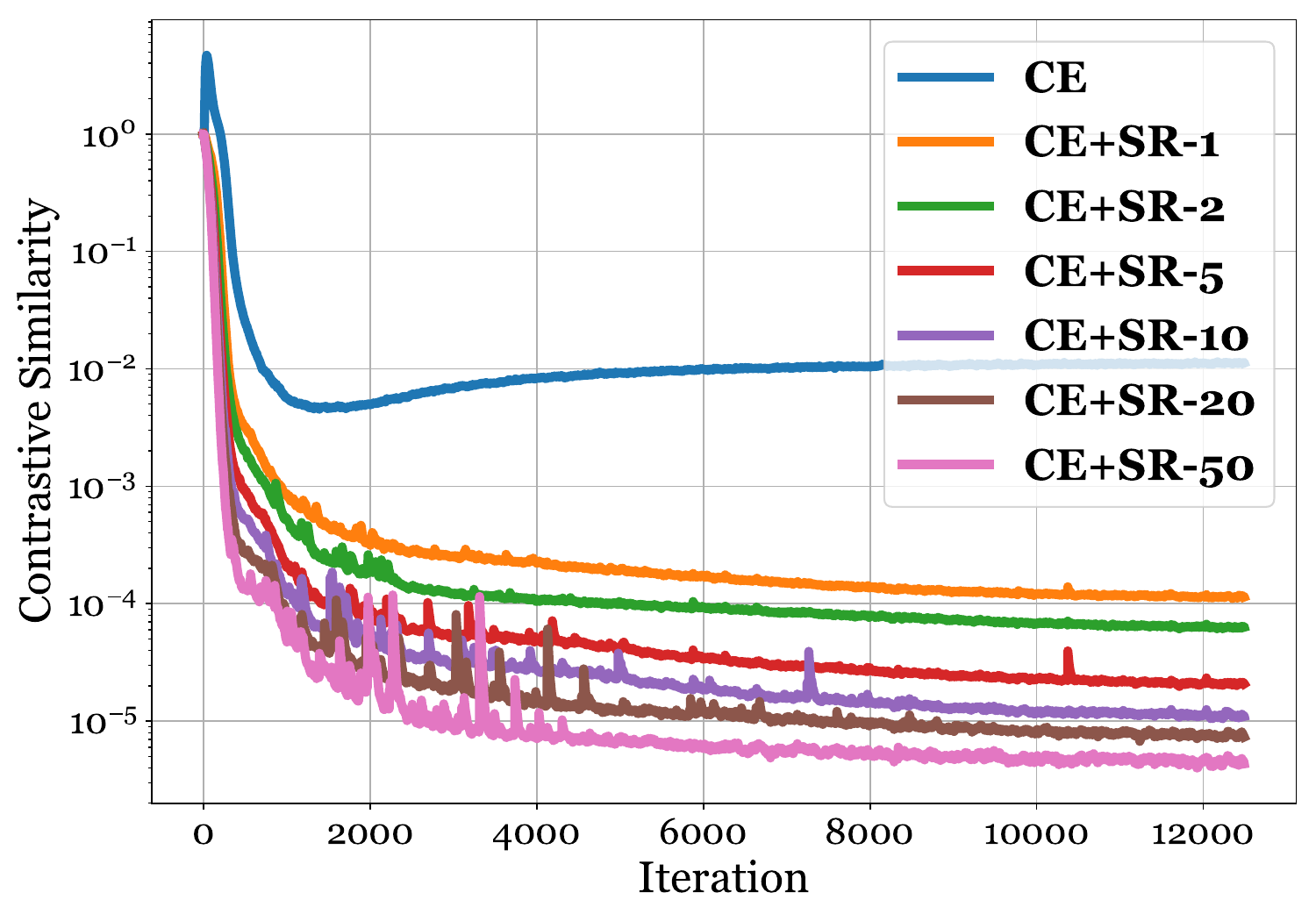}}
\caption{The training curve of the \textsc{SimReg} loss.}
\label{fg:simreg loss curve}
\end{figure}
Figure~\ref{fg:simreg loss curve} illustrates the loss behavior when increasing the coefficient of the \textsc{SimReg} loss. It can be observed that even with a large weighting ratio, \textsc{SimReg} does not cause the training to diverge. At the same time, we also note that the feature consistency loss exhibits a strictly monotonic trend. This phenomenon suggests that \textsc{SimReg} serves as a stable regularization term: rather than interfering with the optimization of cross-entropy, it progressively strengthens the alignment of token embeddings as its weight grows. In practice, this means that a wide range of coefficient values can be applied without destabilizing training, making \textsc{SimReg} highly robust and easy to integrate into large-scale pretraining pipelines.

\textbf{Trade-off of $\lambda$.} Although we generally hope that greater feature separability will lead to better performance, the pretraining process involves not only learning representations but also learning classification. If $\lambda$ is increased without bound, the weight of \textsc{SimReg} may eventually become too dominant and interfere with the optimization of cross-entropy. This phenomenon can be directly observed from the changes in gradient behavior, which provide an intuitive reflection of the trade-off between the two objectives. Table~\ref{tab:ce_sr_tradeoff}  shows the comparison clearly illustrates the effects of cross-entropy and \textsc{SimReg} under different parameter settings.
\begin{table}[H]
\centering
\caption{Changing trend of CrossEntropy and \textsc{SimReg} loss on different $\lambda$.}
\begin{tabular}{c|cccccc}
\toprule
 & $\lambda=1$ & $\lambda=2$ & $\lambda=5$ & $\lambda=10$ & $\lambda=20$ & $\lambda=50$  \\
\midrule
 CrossEntropy    & 2.38 & 2.35 & 2.32 & \textbf{2.30} & 2.33 & 2.35  \\
\midrule
 \textsc{SimReg}~($\times$1e-6) & 104.6 & 56.2 & 18.3 & 9.5 & 6.0 & 3.1 \\
\bottomrule
\end{tabular}
\label{tab:ce_sr_tradeoff}
\end{table}

\newpage
\subsection{Visualization of the Token Embedding Similarity}
\label{ap:visualization}
Here we provide more visualization demos of the true pretraining data samples from C4 dataset on LLaMA-7B.

\textbf{Text1:} [ {\ttfamily so I'm not sure if there's anything holding the back. I do not think there is by wiggling on it but could possibly have a strap or the like.
I would think there must be a way to remove the panel blocking the bottom of the washer. We installed our own washer and used the clips mentioned in the previous post. Here is a PDF file on how they are used and what they look like.
You may want to run your fingers over the entire carpeted lip ... typically, the} ]
\begin{figure}[H]
\centering
	\includegraphics[width=0.9\textwidth]{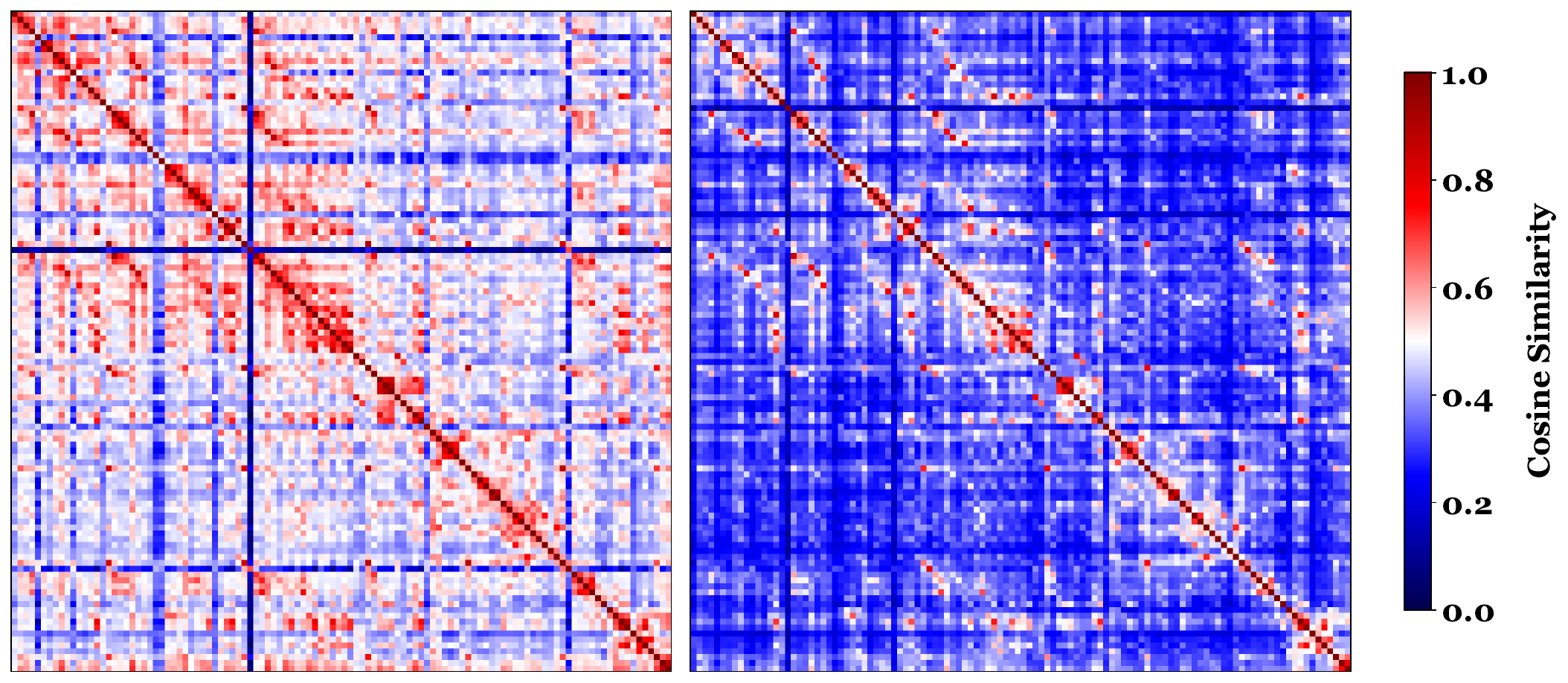}
\caption{The averaged cosine similarity values are 0.488~(CrossEntropy only - left) and 0.354~(CrossEntropy + \textsc{SimReg} - right).}
\end{figure}

\textbf{Text2:} [ {\ttfamily manufacturer runs screws into the floors/cabinets and the heads are buried in the carpet.
There are two screws with square heads in the top of the carpet.
Have you tired to do the recommended procedure to clean the lint out of the drain.
1. Run the unit without clothes and with the dry time off on cycle \# 11.
2. When the water stops entering the unit push and hold the start button until all the lights come on then release the button.} ]
\begin{figure}[H]
\centering
\includegraphics[width=0.9\textwidth]{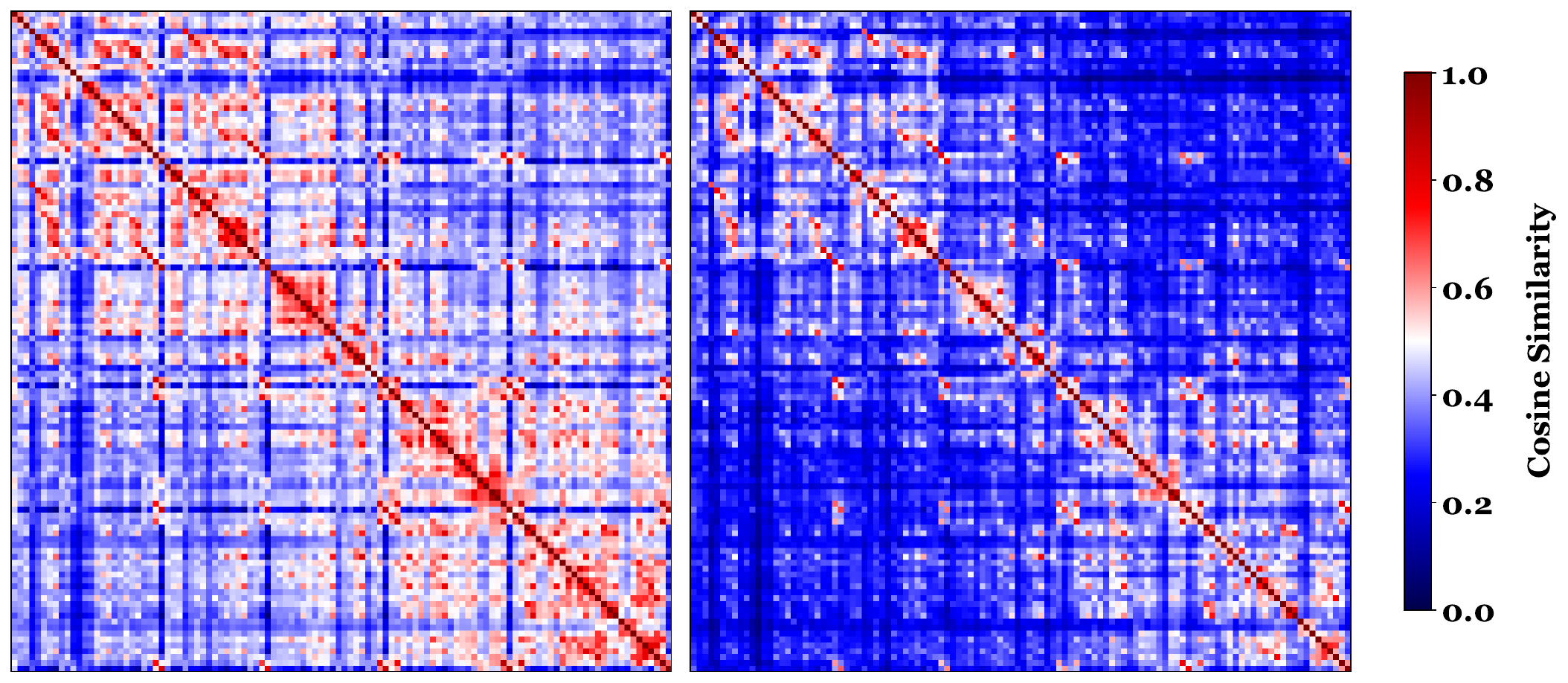}
\caption{The averaged cosine similarity values are 0.445~(CrossEntropy only - left) and 0.333~(CrossEntropy + \textsc{SimReg} - right).}
\end{figure}

\newpage
\section{Appendix: Theoretical Analysis}
\label{ap:proof}
In this section, we mainly demonstrate the theoretical understanding to show how the \textsc{SimReg} loss improves the convergence and generalization efficiency. To this end, we first establish the fundamental properties of the proposed objective and analyze its impact on representation learning. We then present rigorous bounds and intuitive explanations that highlight its advantages over conventional cross-entropy training. These insights not only provide a deeper understanding of why \textsc{SimReg} is effective but also offer useful guidance for its broader application in large-scale pretraining.

\subsection{Relationship between Empirical loss and Margins}
We first introduce the simplified modeling and corresponding notations of the LLM pretraining. Without loss of generality, we decompose the model into two simple parts. The first part is the front-end structure, which takes the raw data as input and outputs the embedding representations. The second part is the back-end structure, which transforms the embeddings into logits, followed by a cross-entropy loss function. We denote $X=\left[x_1, x_2, \cdots, x_n\right]\in\mathbb{R}^{n\times d}$ as the embeddings and $Z=f_{h}(X)\in\mathbb{R}^{n\times c}$ as the logits. The category label are denoted by $Y=\left[y_1,y_2,\cdots,y_n\right]\in\left\{1,2,\cdots,C\right\}^{n}$. For the sample-wise cross entropy loss, we have:
\begin{align*}
    \ell\left(x_i,y_i\right) = -z_{i,y_i} + \log\left(\sum_{j=1}^{K}e^{z_{i,j}}\right).
\end{align*}
The empirical loss is $L=\frac{1}{n}\sum_{i=1}^{n}\ell(x_i,y_i)$. Then we consider the margin value in multi-class classification, which is also the joint gaps of different categories $m_i = z_{i,y_i} - \max_{j\neq y_i} z_{i,j}$. Therefore, we have:
\begin{align*}
    \ell\left(x_i,y_i\right) = \log\left(1+\sum_{j\neq y_i}e^{-\left(z_{i,y_i}-z_{i,j}\right)}\right)\leq\log\left(1+\left(C-1\right)e^{-m_i}\right) \leq \left(C-1\right)e^{-m_i},
\end{align*}
where the empirical loss is $L=\frac{1}{n}\sum_{i=1}^{n}\ell(x_i,y_i)\leq\frac{C-1}{n}\sum_{i=1}^n e^{-m_i}$. Generally, if the classification margins of all samples are increased by at least $\Delta \geq 0$, the loss will be multiplicatively reduced by a factor of $e^{-\Delta}$.

\subsection{Equivalent Constraint of the SimReg Loss}
Here we learn how the \textsc{SimReg} loss affect the embeddings and the model performance. Here we let each embedding $\mathbf{e}_i=r_i\mathbf{a}_i$ where $r_i=\Vert \mathbf{e}_i \Vert\geq 0$ is the magnitude and $\mathbf{a}_i$ is the normalized embedding. \textsc{SimReg} loss evaluates the exponential of the cosine similarity of two embeddings. Its core focus lies in the geometric information of the term $a$. To learn the performance of the \textsc{SimReg}, for each label $y_i$, we define a positive set $\mathcal{P}_i=\left\{a_j:y_j=y_i\right\}$ and a negative set $\mathcal{N}_i=\left\{a_j:y_j\neq y_i\right\}$. The union of $\mathcal{P}_i$ and $\mathcal{N}_i$ always combines a complete sequence. 

To understand the performance of \textsc{SimReg} in detail, we first introduce a general kernal function $\kappa\left(\mathbf{u},\mathbf{v}\right)=\exp\left(\mathbf{u}^\top \mathbf{v}\right)$, which admits the Maclaurin series $\kappa\left(\mathbf{u},\mathbf{v}\right)=\sum_{m=0}^{\infty}\frac{\left(\mathbf{u}^\top \mathbf{v}\right)^m}{m!}$.
It is a positive definite kernel on the unit sphere. By introducing an explicit map: $h:\mathbb{S}^{d-1}\rightarrow \mathcal{H}$ on the symmetric tensor powers:
\begin{equation}
    h(\mathbf{u})=\left[1,\frac{1}{\sqrt{\pi}}\mathbf{u}, \frac{1}{\sqrt{2!\pi^2}}\text{vec}\left(\mathbf{u}^{\otimes 2}\right), \frac{1}{\sqrt{3!\pi^3}}\text{vec}\left(\mathbf{u}^{\otimes 3}\right),\cdots\right],
\end{equation}
thus we have the transformation of $\langle h(\mathbf{u}), h(\mathbf{v})\rangle=\kappa\left(\mathbf{u},\mathbf{v}\right)$. The mapping $h$ is to construct a linear expansion of $\kappa$ in the reproducing kernel Hilbert space~(RKHS) $\mathcal{H}$. Therefore, we have:
\begin{align*}
    \log\left(\sum_{i\in\mathcal{P}_k}\exp\left(\mathbf{e}_k^\top\mathbf{e}_i\right)\right) = \log\left(\sum_{i\in\mathcal{P}_k}\left\langle h(\mathbf{e}_k), h(\mathbf{e}_i) \right\rangle\right) = \log\left(\left\langle h(\mathbf{e}_k),\mu_k^+\right\rangle\right) + \log\left(\vert\mathcal{P}_k\vert\right),
\end{align*}
where $\mu_k^+=\frac{1}{\vert\mathcal{P}_k\vert}\sum_{i\in\mathcal{P}_k}h(\mathbf{e}_i)$ is the positive kernel means. Here $\vert\mathcal{P}_k\vert$ can be considered as a offset to scale the positive samples. The theoretical analysis can be symmetrically extended to negative samples, yielding an equivalent conclusion.

Therefore, the \textsc{SimReg} loss consider the difference between teh positive and negative set by:
\begin{align*}
    \min_{\mathbf{e}=f_E(\mathbf{x})} \ J = \mathbb{E}_{\mathbf{x}} \log\left(\frac{\left\langle h(\mathbf{e}_k),\mu_k^-\right\rangle}{\left\langle h(\mathbf{e}_k),\mu_k^+\right\rangle}\right) + \log\left(\frac{\vert\mathcal{N}_k\vert}{\vert\mathcal{P}_k\vert}\right).
\end{align*}
The ratio of positive to negative samples only affects the scale of the loss, but does not alter the primary optimization objective of the first term. It pushes the anchor direction to align with the positive kernel mean and to anti-align with the negative kernel mean. It also nudges the group means themselves: positives move toward anchors that they are already close to, and negatives move away in the RKHS sense. We also have the nearest positive prototype for each class:
\begin{align*}
    \max_{\Vert \mathbf{e}\Vert}\left\langle h(\mathbf{e}),\mu_k^+\right\rangle = \Vert h(\mathbf{e})\Vert \Vert \mu_k^+ \Vert = \kappa(\mathbf{e},\mathbf{e})\Vert\mu_k^+\Vert = \sqrt{e}\Vert\mu_k^+\Vert.
\end{align*}
The same, the $\sqrt{e}$ scaling also hold for the negative set. Beyond the optimization objective itself, we can further consider the problem from the perspective of gradient directions to refine the learning target. By considering the Fr\'echet gradient, we have:
\begin{align*}
    \nabla_{h(\mathbf{e}_k)} J = \frac{\mu_k^-}{\left\langle h(\mathbf{e}_k),\mu_k^-\right\rangle} - \frac{\mu_k^+}{\left\langle h(\mathbf{e}_k),\mu_k^+\right\rangle}.
\end{align*}
Generally, $\mu_k^-\neq\mu_k^+$. From the gradient expression, we can see that the optimization dynamics naturally combine both “attractive” and “repulsive” effects. Specifically, the first term pushes the representation $h(\mathbf{e}_k)$ away from the negative center $\mu_k^-$, while the second term pulls it closer to the positive center $\mu_k^+$. As a result, the overall update direction is shaped by the joint effect of being attracted to positives and repelled from negatives, thereby optimizing the representation space effectively. From the above two perspectives, it is clear that \textsc{SimReg} enforces feature consistency alignment in the RKHS sense.

\subsection{Center-aligned Embeddings Can Enhance Optimization}
Then we consider the performance of the center-aligned embedding. To learn the transferred impact from the mapping $h(\mathbf{e}_k)$ to vanilla variable $\mathbf{e}_k$, we first consider the normalized $\mathbf{a}_k$ term, where the cosine similarity can be considered as $\mathbf{a}_k^\top\mathbf{a}_j$. To simplify the notation, we additionally define the weighted average direction of a variable $\mathbf{a}$ over its associated positive and negative sets by $\mathbf{v}_k^+=\frac{1}{\Vert \mathcal{P}_k\Vert}\sum_{i\in\mathcal{P}_k}\exp\left(\mathbf{a}_k^\top\mathbf{a}_i\right)\mathbf{a}_{i}$ and $\mathbf{v}_k^-=\frac{1}{\Vert \mathcal{N}_k\Vert}\sum_{j\in\mathcal{N}_k}\exp\left(\mathbf{a}_k^\top\mathbf{a}_j\right)\mathbf{a}_{j}$. Similarly, we also define the loss of positive set and negative set as $P_k$ and $N_k$. Therefore, we have the following gradient form:
\begin{align*}
    \nabla_{\mathbf{a}_k}L_{\text{sr}} = \frac{N_k}{P_k + N_k}\left(\mathbf{v}_k^- - \mathbf{v}_k^+ \right).
\end{align*}
Since the $\mathbf{a}_k$ is constrainted by $\Vert\mathbf{a}_k\Vert=1$, the true update direction is obtained by projecting the gradient onto the tangent space: $-\prod_{\mathbf{v}_k}\nabla_{\mathbf{a}_k}L_{\text{sr}}=-\frac{N_k}{P_k + N_k}\left(I-\mathbf{a}_k\mathbf{a}_k^\top\right)\left(\mathbf{v}_k^- - \mathbf{v}_k^+ \right)$. Next, we analyze how the gradient dynamics associated with the positive sample set vary along the update direction. This dynamic essentially characterizes how strongly the representation is pulled toward the positive center during optimization. A larger value indicates that the update direction aligns well with the attraction force from positive samples, thereby accelerating convergence. Conversely, a smaller value reflects weaker alignment, suggesting limited contribution from positive samples in shaping the optimization trajectory. For the positive sample loss, we obtain~(for clarity of exposition, we omit constant scalar terms):
\begin{align*}
    \frac{d}{dt}\Vert\mathbf{a}_k-\mathbf{v}_k^+\Vert^2 
    &= 2\left(\mathbf{a}_k-\mathbf{v}_k^+\right)^\top\left(I-\mathbf{a}_k\mathbf{a}_k^\top\right)\mathbf{v}_k^+ - \underbrace{2\left(\mathbf{a}_k-\mathbf{v}_k^+\right)^\top\left(I-\mathbf{a}_k\mathbf{a}_k^\top\right)\mathbf{v}_k^-}_{\text{negative perturbation}}.
\end{align*}
When treating the update on the negative sample set as a small perturbation to that on the positive samples, we have $\frac{d}{dt}\Vert\mathbf{a}_k-\mathbf{v}_k^+\Vert^2\leq 2\left(\mathbf{a}_k^\top\mathbf{v}_k^+\right)^2-\Vert\mathbf{v}_k^+\Vert^2 \leq 0$. Similarly, the gradient dynamics on the negative sample set can be obtained as $\frac{d}{dt}\Vert\mathbf{a}_k-\mathbf{v}_k^-\Vert^2\geq 0$. In conclusion, taking a small step along the tangent update direction inherently drives the representation closer to the weighted center of the positive class while simultaneously pushing it away from that of the negative class. In other words, such updates reinforce the consistency among positive samples and reduce the influence of negatives, thereby shaping a clearer separation in the feature space. Importantly, this property does not rely on any assumptions about the underlying functional form, but rather arises directly from the optimization objective itself, ensuring both generality and robustness. To further refine the update dynamics, a temperature coefficient can be introduced as a scaling factor. By adjusting the sharpness of the similarity distribution, the temperature effectively controls the relative strength of attraction toward positive samples and repulsion from negative samples. In particular, incorporating a temperature into the formulation normalizes the gradient magnitudes and ensures that the update direction satisfies the desired balance condition between positive and negative contributions. This modification not only stabilizes training but also enhances the flexibility of the loss function in adapting to different representation scales. This result can be directly extended from the normalized variables to the original embedding variables $\mathbf{e}$, thereby completing the proofs.
%%%%%%%%%%%%%%%%%%%%%%%%%%%%%%%%%%%%%%%%%%%%%%%%%%%%%%%%%%%%%%%%%%%%%%%%%%%%%%%
%%%%%%%%%%%%%%%%%%%%%%%%%%%%%%%%%%%%%%%%%%%%%%%%%%%%%%%%%%%%%%%%%%%%%%%%%%%%%%%
%%%%%%%%%%%%%%%%%%%%%%%%%%%%%%%%%%%%%%%%%%%%%%%%%%%%%%%%%%%%

% \newpage
% \input{texts/checklist}

\end{document}